\documentclass[10pt,twocolumn]{article}
\usepackage{times}
\usepackage{graphicx}
\usepackage{booktabs} 
\usepackage{amsmath}
\usepackage{amsfonts}
\usepackage{comment}
\usepackage{enumerate}
\usepackage{lastpage}
\usepackage{color}
\usepackage{caption}
\usepackage{subcaption}
\usepackage[paper]{optional} 
\usepackage{url}

\newcommand{\chuan}[1]{\textcolor{red}{{\bf Chuan:} #1}}

\def\eg{\textit{e.g.}}
\def\ie{\textit{i.e.}}
\def\DL2{{DL}$^2$}

\begin{document}

\title{\Large DL$^2$: A Deep Learning-driven Scheduler for Deep Learning Clusters}
\author{Yanghua Peng$^1$ \quad Yixin Bao$^1$ \quad Yangrui Chen$^1$ \quad Chuan Wu$^1$ \quad Chen Meng$^2$ \quad Wei Lin$^2$\\[2mm]
\small {$^1$\{yhpeng, yxbao, yrchen, cwu\}@cs.hku.hk \quad
          $^2$\{mc119496, weilin.lw\}@alibaba-inc.com} \\ [2mm]
\small {\em  $^1$The University of Hong Kong \quad
          $^2$Alibaba Inc.} \\ [2mm]
}
\date{}
\maketitle

\begin{abstract}
More and more companies have deployed machine learning (ML) clusters, where deep learning (DL) models are trained for providing various AI-driven services. Efficient resource scheduling is essential for maximal utilization of expensive DL clusters. Existing cluster schedulers either are agnostic to ML workload characteristics, or use scheduling heuristics based on operators' understanding of particular ML framework and workload, which are less efficient or not general enough. In this paper, we show that DL techniques can be adopted to design a generic and efficient scheduler.

\DL2~is a DL-driven scheduler for DL clusters, targeting global training job expedition by dynamically resizing resources allocated to jobs. \DL2~advocates a joint supervised learning and reinforcement learning approach: a neural network is warmed up via offline supervised learning based on job traces produced by the existing cluster scheduler; then the neural network is plugged into the live DL cluster, fine-tuned by reinforcement learning carried out throughout the training progress of the DL jobs, and used for deciding job resource allocation in an online fashion. By applying past decisions made by the existing cluster scheduler in the preparatory supervised learning phase, our approach enables a smooth transition from existing scheduler, and renders a high-quality scheduler in minimizing average training completion time. We implement \DL2~on Kubernetes and enable dynamic resource scaling in DL jobs on MXNet. Extensive evaluation shows that \DL2~outperforms fairness scheduler (\ie, DRF) by $44.1\%$ and expert heuristic scheduler (\ie, Optimus) by $17.5\%$ in terms of average job completion time.
\end{abstract}

\section{Introduction}
Recent years have witnessed the breakthrough of DL-based techniques in various domains, such as machine translation~\cite{bahdanau2015neural}, image classification~\cite{krizhevsky2012imagenet} and speech recognition~\cite{graves2013speech}. Large companies have deployed ML clusters with tens to thousands of expensive GPU servers, and run distributed training jobs on one or different distributed ML frameworks (such as TensorFlow~\cite{abadi2016tensorflow}, MXNet~\cite{chen2016mxnet}, Petuum~\cite{xing2015petuum} and PaddlePaddle~\cite{PaddlePaddle}), to obtain DL models in need for their AI-driven services. Even with parallel training, training a DL model is commonly very time and resource intensive. Efficient resource scheduling is crucial in operating a shared DL cluster with multiple training jobs, for best utilization of expensive resources and expedited training completion.

Two camps of schedulers exist in today's ML clusters. In the first camp, general-purpose cloud/cluster schedulers are applied, and possibly customized, for distributed ML job scheduling. For example, Google uses Borg~\cite{verma2015large} as its DL cluster scheduler; Microsoft, Tencent, and Baidu use custom versions of YARN-like schedulers~\cite{vavilapalli2013apache} for managing DL jobs. Representative scheduling strategies used include First-In-First-Out (FIFO) 
and Dominant Resource Fairness (DRF)~\cite{ghodsi2011dominant}. 
These schedulers allocate resources according to user specification and do not adjust resource allocation during training. As we will see in \S\ref{subsec:motivations}, setting the right amount of resources for a job is difficult and static resource allocation leads to resource under-utilization in the cluster.
 
In the second camp, recent studies have proposed white-box heuristics for resource allocation in ML clusters~\cite{zhang2017slaq}\cite{peng2018optimus}\cite{bao2018online}. Typically they tackle the problem in two steps: set up analytical models for DL/ML workloads, and propose scheduling heuristics accordingly for online resource allocation and adjustment. Designing heuristics requires a deep understanding of ML frameworks and workloads, and the analytical model is tightly coupled with the ML framework implementation (\eg, a new feature or optimization in evolving ML frameworks may invalidate the analytical model)~\cite{peng2018optimus}. Further, the modeling typically does not consider interference in a multi-tenant cluster, where in average $27.3\%$ performance variation may happen (\S\ref{subsec:motivations}).

In this paper, we pursue a DL cluster scheduler that does not depend on expert heuristics and explicit modeling, resorting to a black-box end-to-end approach enabled by modern learning techniques. We propose \DL2, a deep learning-driven scheduler for deep learning clusters, that dynamically adjusts resource allocation to training jobs on the go. \DL2~learns resource allocation policies through experience using deep reinforcement learning (DRL): the policy neural network takes the current system state as input, produces resource allocation decisions for all the current training jobs and gradually improves the decisions based on feedback. However, merely applying off-the-shelf RL algorithms to scheduling does not produce high-quality decisions, and careful design according to the problem nature is in need.

Existing DRL applications in resource scheduling scenarios \cite{mao2016resource}\cite{mao2018learning} \cite{mao2019learning} (\S\ref{related_DL_in_network}) use simulators to generate training data for offline training, and apply trained models for resource scheduling in a live system. The core of such a simulator is typically an explicit performance model as mentioned above, and hence the inaccuracy of the simulator may lead to low-quality trained model. Instead of extensive offline training over large simulation, \DL2~takes a different approach: we bootstrap the model using minimal offline supervised learning with any available historical job traces and decisions of any existing scheduling strategy employed in the cluster; then we use online training with feedback from ongoing decision making in a live system, with carefully designed techniques to guide model convergence to high-quality decisions, which minimize average job completion time in the cluster.

In summary, we make the following contributions in \DL2:

$\triangleright$ In contrast to previous DL cluster scheduling approaches that require analytical performance model and job profiling, \DL2~adopts a more generic design, \ie, using DRL to schedule DL workloads. Instead of simulation-driven RL model training, we adopt online training with real feedback from online resource allocation (\S\ref{sec:background}).

$\triangleright$ We identify that direct application of simple RL approaches for our online scheduler training often leads to poor decisions. 
 To avoid 
poor decisions at the beginning of online RL, we apply past decisions made by an existing scheduler in the DL cluster in a preparatory offline supervised learning stage. Our approach enables a smooth transition from an existing scheduler, and automatically learns a better scheduler beyond the performance level of the existing one (\S\ref{sec:overview}). To optimize online RL particularly for DL job scheduling, 
 we propose job-aware exploration for efficient exploration in the action space, and adopt additional training techniques (\eg, actor-critic algorithm, experience replay) for sample-efficient learning (\S\ref{sec:joint_learning}).


$\triangleright$ We design and implement elastic scaling in a popular distributed ML framework, MXNet~\cite{chen2016mxnet}, to achieve dynamic worker/parameter server adjustment (\S\ref{sec:implementation}). We integrate \DL2~with Kubernetes~\cite{kubernetes}, 
and carefully evaluate \DL2~using testbed experiments and controlled simulations, driven by DL job patterns collected from a production DL cluster. 
Evaluation results show that \DL2~significantly outperforms other representative schedulers in various scenarios, \eg, $44.1\%$ improvement in average job completion time as compared to the widely adopted DRF scheduler. We also demonstrate \DL2's scaling overhead and generality (\S\ref{sec:evaluation}).

\section{Background and Motivation}
\label{sec:background}

\subsection{The Parameter Server Framework}

We focus on the parameter server (PS) architecture~\cite{li2014scaling}, which is widely adopted in distributed ML learning frameworks for parallel training, such as in MXNet~\cite{chen2016mxnet}, TensorFlow~\cite{abadi2016tensorflow}, PaddlePaddle~\cite{PaddlePaddle} and Angel~\cite{Tencent_Angel}. Note that \DL2~can also be extended to all-reduce, as discussed in \S\ref{sec:discussions}. In the PS architecture, the model, \eg, a deep neural network (DNN), is partitioned among multiple parameter servers (PSs) and training dataset is split among workers (\ie, in the representative data parallel training model). The data partition at each worker is divided into mini-batches; each worker processes a mini-batch locally and computes model parameter changes, typically expressed as gradients. The gradients are pushed to PSs which maintain global model parameters. 
 We focus on synchronous training, where the workers' training progresses are synchronized and PSs update the global model after receiving gradients from all workers in each iteration. 
 Updated parameters are sent back to the workers. A worker starts the next training iteration/step by processing the next mini-batch with the updated parameters. After all mini-batches in the entire dataset have been processed once, one {\em training epoch} is done. The input dataset is usually trained for multiple epochs until the model converges.

\begin{figure}
    \centering
    \begin{minipage}[t]{0.48\linewidth}
   	\centering
        \includegraphics[width=\linewidth]{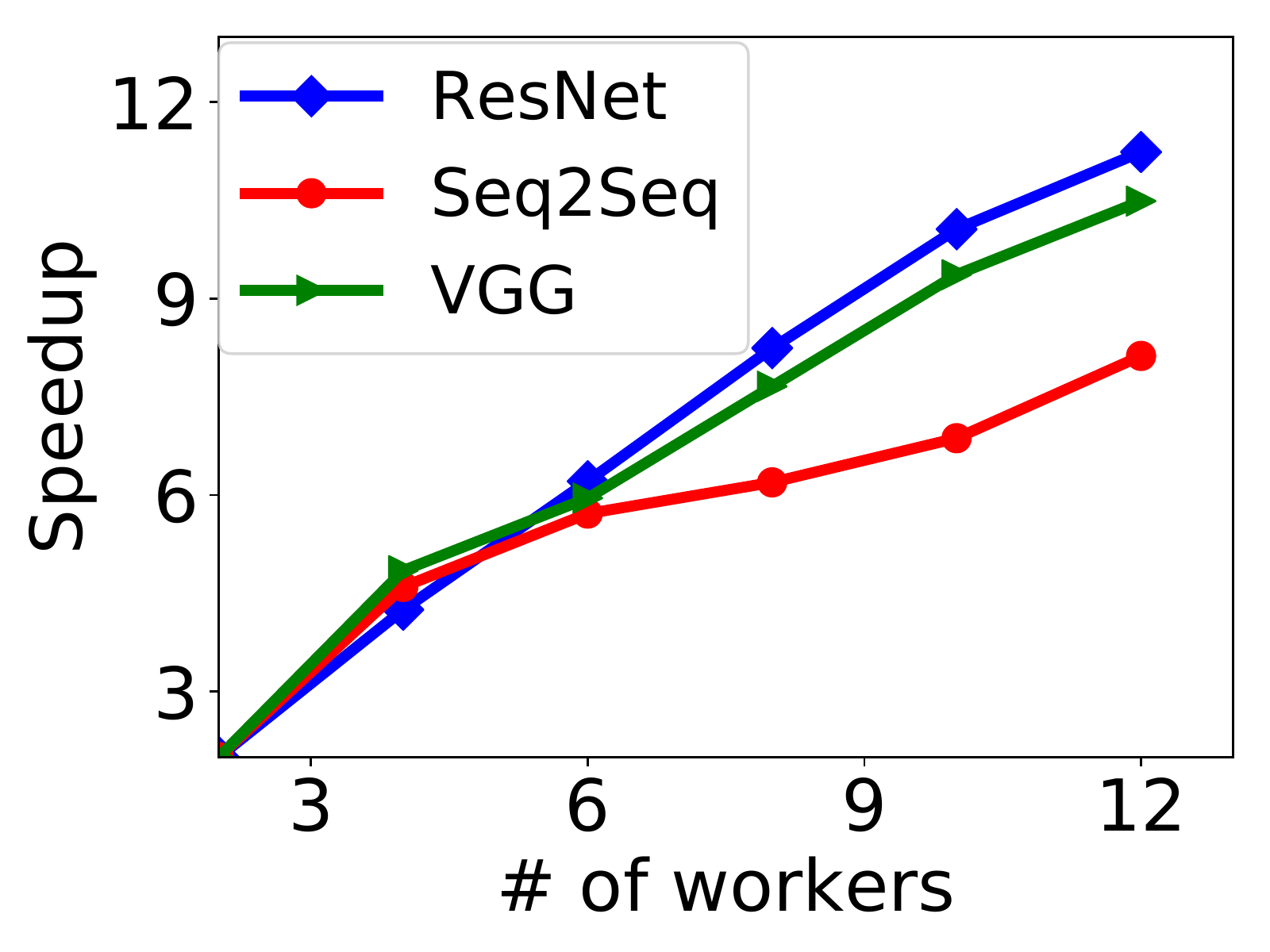}
        \caption{Training speedup with diff. worker/PS numbers}
        \label{fig:scalability}
    \end{minipage}
    \hfill 
   \begin{minipage}[t]{0.48\linewidth}
    \centering
        \includegraphics[width=\linewidth]{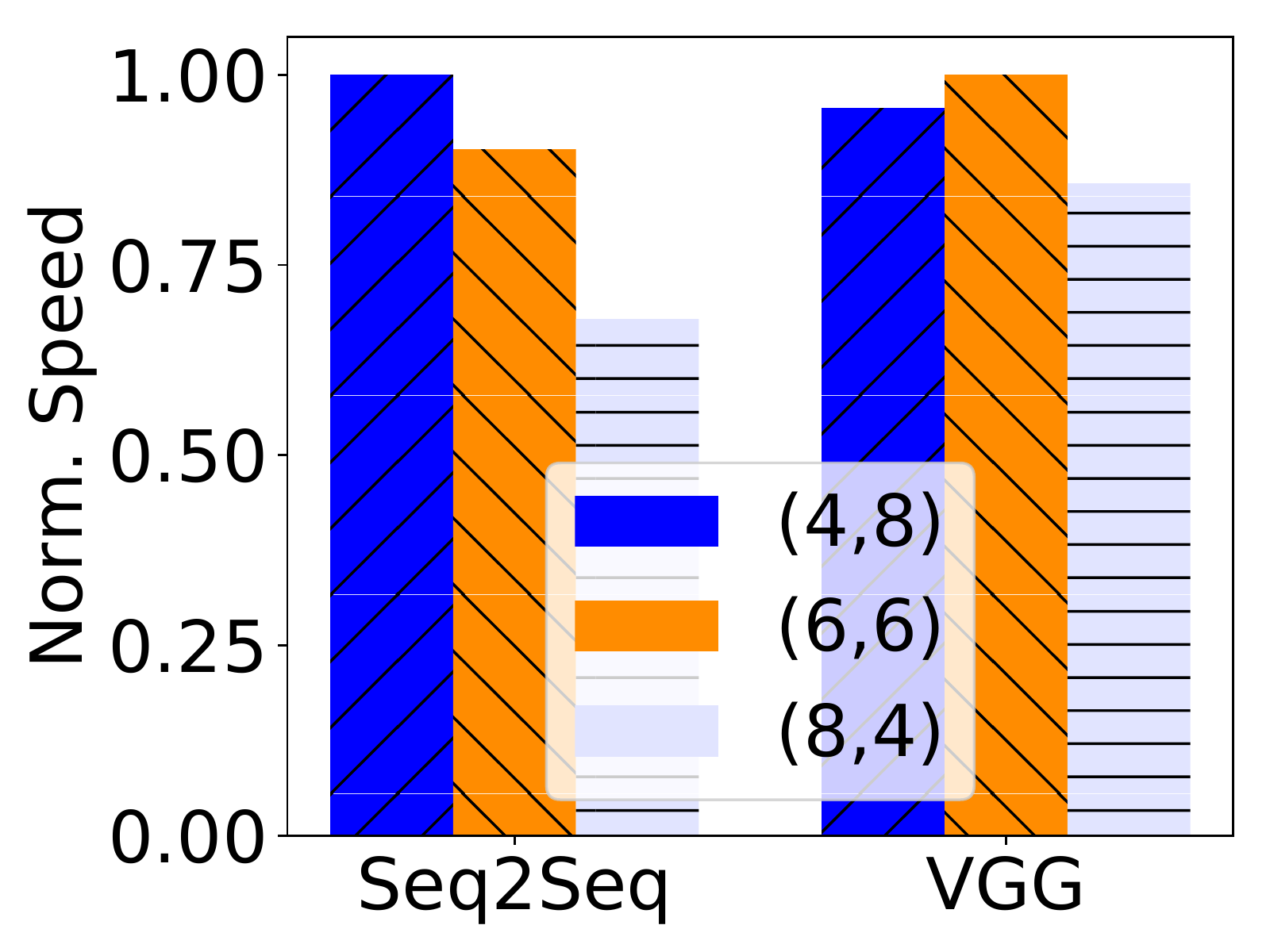}
        \caption{Training speed under diff. PS-to-worker ratios}
       \label{fig:ps_worker_ratio}
    \end{minipage}
\vspace{-5mm}
\end{figure}

\subsection{Motivations}
\label{subsec:motivations}
The typical workflow for a user to train a model in a DL cluster is as follows: The user specifies how many PSs and workers she/he wishes to use and the amount of resources (\eg, GPU, CPU) each PS/worker needs, and then submits the job to the scheduler (\eg, Borg~\cite{verma2015large}, YARN~\cite{vavilapalli2013apache}, Mesos~\cite{hindman2011mesos}). The scheduler allocates PSs and workers to the job according to both user demand and its scheduling strategy, and the allocated resources then remain fixed over the entire training course of the job. This workflow has two limitations, as illustrated below.

\noindent\textbf{Difficulty in setting the right worker/PS numbers.} How does a job's training speed improve when more PSs and workers are added to the job? We train 3 classical models, \ie, ResNet-50~\cite{he2016deep}, VGG-16~\cite{simonyan2014very} and Seq2Seq~\cite{gehring2017convolutional}, in our testbed of 6 machines (see \S\ref{sec:evaluation} for hardware details), and measure their training speeds (in terms of the number of samples trained per unit time), when increasing the number of workers and keeping the number of PSs equal to the worker number. Each worker uses 1 GPU, 4 CPU cores, 10GB memory and each PS has 4 CPU cores, 10GB memory. In Fig.~\ref{fig:scalability}, the speedup is calculated by dividing the training speed achieved using multiple workers/PSs (they are deployed onto physical machines in a load-balanced fashion) by the training speed obtained using one worker and one PS colocated on a single machine. We observe a trend of decreasing return, \ie, adding PSs/workers does not improve the training speed linearly. This is because more communication overhead is incurred when there are more PSs or workers. 

On the other hand, is an equal number of PSs and workers (as a general rule of thumb) always the best? We fix the total number of PSs and workers to be 12 and measure the training speed of two models under different combinations of PS/worker numbers (\ie, 4:8, 6:6, 8:4)~\cite{peng2018optimus}. 
 Fig.~\ref{fig:ps_worker_ratio} shows that 
 Seq2Seq achieves highest training speed when there are 4 PSs and 8 workers, while VGG-16 is trained fastest with 6 PSs and 6 workers. 

From the above, we see that it is challenging to reason about which job will have the largest marginal gain from extra resources and 
what the best PS-to-worker ratio is, as they are affected by many factors, \eg, allocated resources, models. Existing schedulers largely side-step this problem and leave it to the user to decide how many PSs/workers to use.

\begin{figure}
    \centering
    \begin{minipage}[t]{0.48\linewidth}
    \centering
        \includegraphics[width=\linewidth]{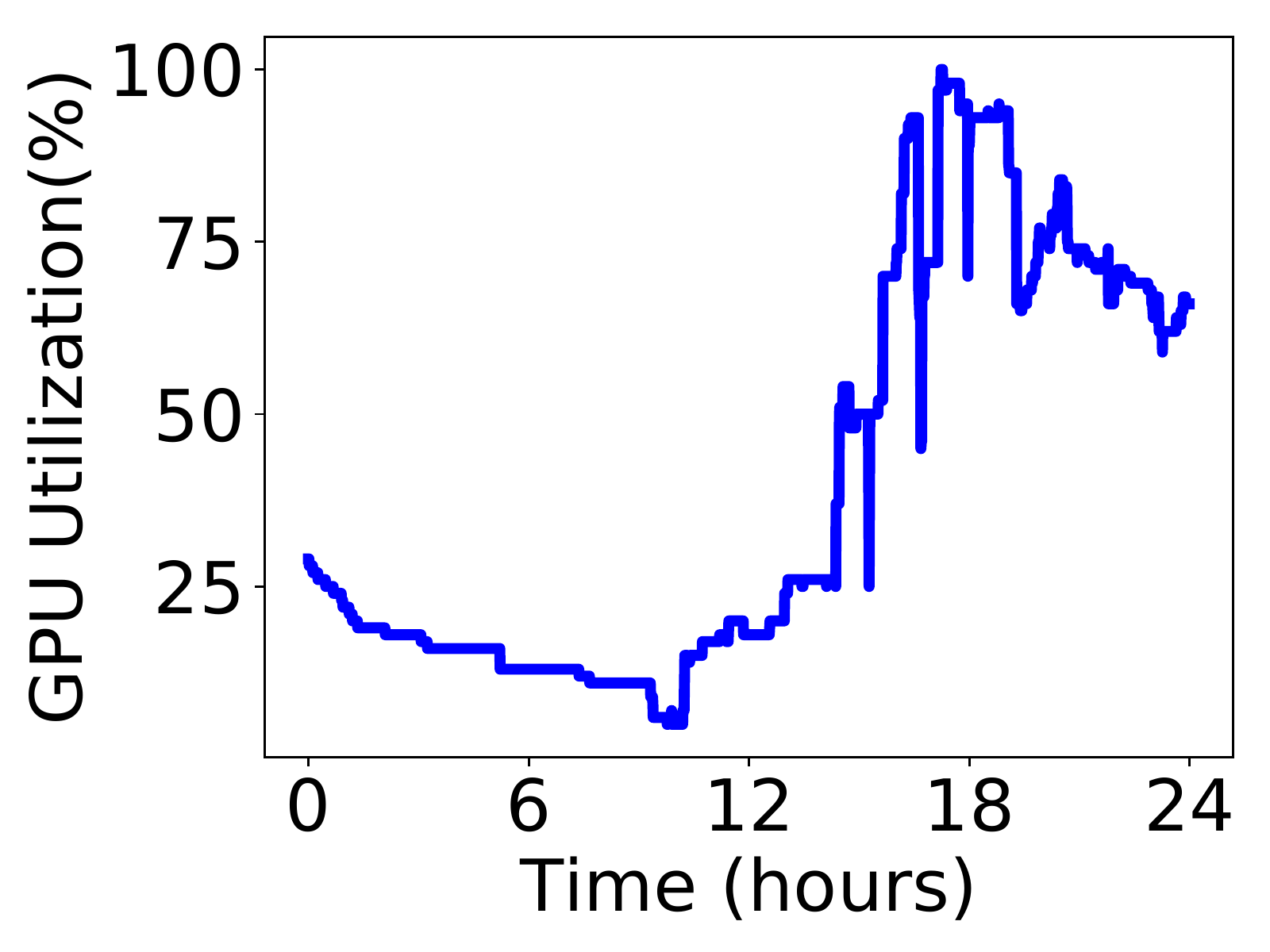}
        \caption{GPU utilization in a production DL cluster}
       \label{fig:gpu_utilization}
    \end{minipage}
    \hfill 
    \begin{minipage}[t]{0.48\linewidth}
   	\centering
        \includegraphics[width=\linewidth]{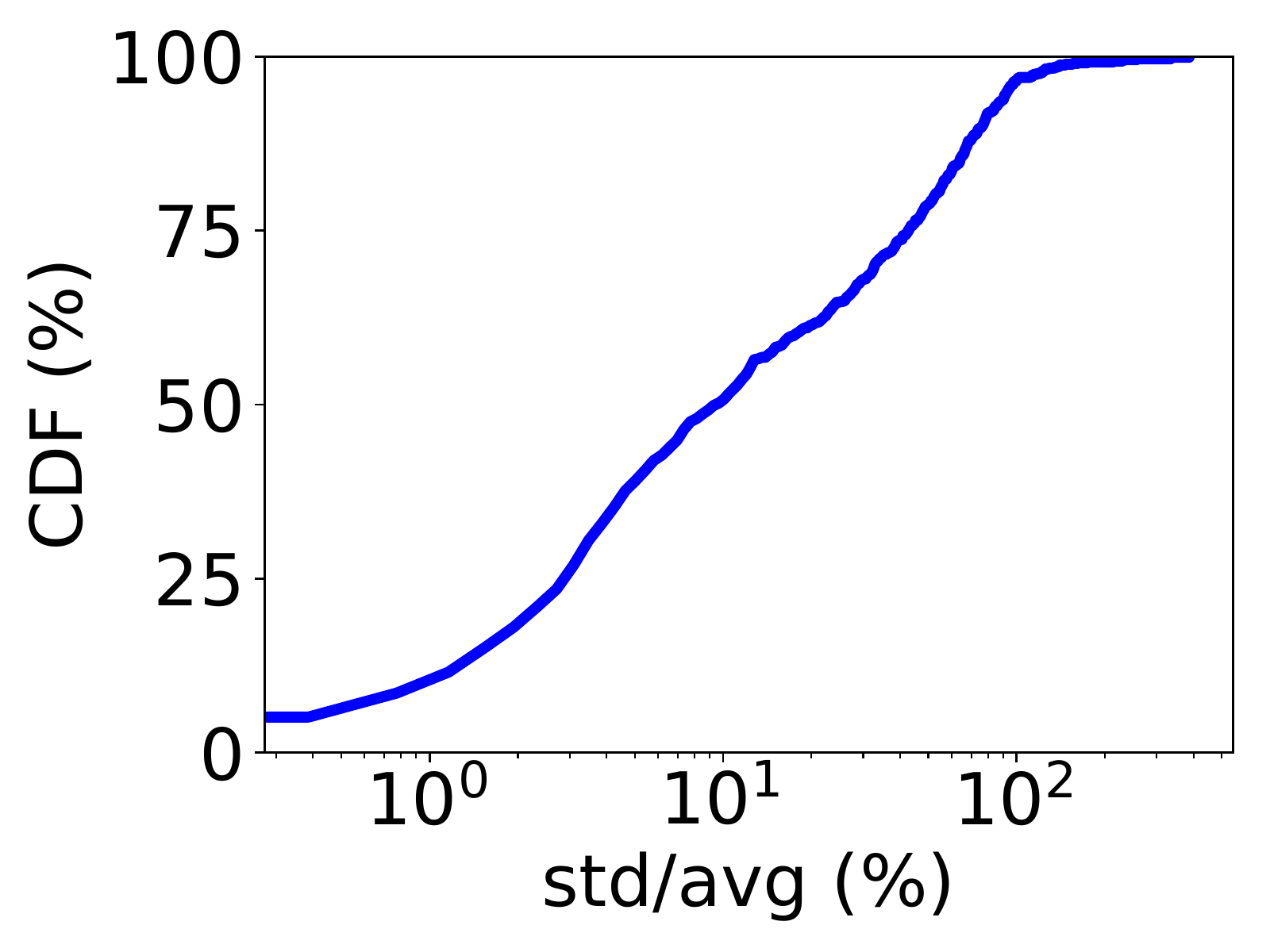}
        \caption{Variation of training completion time}
        \label{fig:interference}
    \end{minipage}
\vspace{-5mm}
\end{figure}

\noindent\textbf{Static resource allocation.} The GPU cluster resources are often not fully utilized: when a training job is completed, the resources it releases (\eg, expensive GPUs) may become idle, rather than being exploited by remaining jobs that are still running. Fig.~\ref{fig:gpu_utilization} shows the GPU utilization during a 24-hour interval in a production DL cluster with about 1000 P100 GPU cards (company name removed due to anonymity requirement), whose job traces will be used in our evaluation (\S\ref{sec:evaluation}). We see that the GPU utilization level varies significantly over time, providing opportunity for dynamic resource scaling out/in in training jobs when cluster load is low/high. 

We advocate dynamic adjustment of worker/PS numbers in training jobs over time, to maximally utilize available resources in the DL cluster to expedite job completion. With this, we further do not require users to submit the number of workers/PSs they want to use in their jobs (who nonetheless may not be at the best position to decide that), but will decide the best worker/PS numbers for each user at each time based on both global resource availability and individual jobs' performance.


\noindent\textbf{White-box heuristics.} There have been existing studies which explicitly model detailed relationship between the training speed and resources within jobs, and design scheduling heuristics based on the resource-speed model, \eg, SLAQ~\cite{zhang2017slaq}, Optimus~\cite{peng2018optimus} and OASiS~\cite{bao2018online}. They have two limitations. {\em First}, in order to derive an accurate performance model, the modeling process is coupled tightly with ML framework implementation, and re-modeling is often needed when the framework changes (\eg, adding new features or adopting optimization). For example, Optimus models computation and communication as two separate procedures during one training step; its model needs to be rebuilt when new features are incorporated into ML frameworks, \eg, overlapping backward computation with communication, gradient compression~\cite{chen2016mxnet}.

{\em Second}, explicit performance models are built without considering interference in multi-tenant GPU clusters. For example, SLAQ~\cite{zhang2017slaq} and Optimus~\cite{peng2018optimus} assume no network congestion on PSs, and OASiS~\cite{bao2018online} and Optimus~\cite{peng2018optimus} assume that the available bandwidth is a constant. However, we observe that the speed for training the same model may vary significantly. Fig.~\ref{fig:interference} shows the performance variation (\ie, the standard deviation of completion time of a training job divided by average completion time of the job over its multiple runs) of 898 DL jobs from the production ML cluster trace. The average variation is $27.3\%$ and the variation for some jobs ($3.5\%$ of all jobs) even exceeds $100\%$. Besides, explicitly modeling interference among ML jobs is also very difficult~\cite{bao2019deep}, as each additional dimension (neural network structure, parallelism architecture, runtime isolation, etc.) increases complexity.

In contrast to white-box model-based schedulers, we resort to a black-box approach and design an RL-based resource scheduler: it automatically learns end-to-end resource allocation policy without requiring expert heuristics and without explicitly modeling the ML framework, the workload, and the interference.

\subsection{Deep Reinforcement Learning}

DRL has been widely used for sequential decision making in an unknown environment, where the agent learns a policy to optimize a cumulative reward by trial-and-error interactions with the environment~\cite{sutton1998reinforcement}. In each iteration, the agent observes the current state of the environment and then chooses an action based on the current policy. The environment moves to a new state and reveals the reward, and the policy is updated based on the received reward.

Existing DRL-based schedulers for resource allocation~\cite{mao2016resource}\cite{chen2017deep}\cite{mao2018learning}\cite{mao2019learning} generate a large amount of traces for offline DRL model training, typically by building an explicit resource-performance model for jobs and using it to estimate job progress based on the allocated resources, in the offline simulation environment. The need for model rebuilding (due to ML system changes) and inaccuracy (due to interference) of the performance model degrade the quality of the DRL policy learned (see Fig.~\ref{fig:performance_comparison}). Another possibility is to use available historical traces for offline DRL training. However, due to the large decision space of resource allocation (exponential with the amount of resources), historical traces usually do not include feedback for all possible decisions produced by the DRL policy~\cite{mao2016resource}\cite{mao2018learning}\cite{bao2019deep}.

Therefore, instead of offline training in a simulated environment, we advocate online RL in the live cluster, exploiting true feedback for resource allocation decisions produced by the DRL agent, to learn a good policy over time. Pure online learning of the policy network model from scratch can result in poor policies at the beginning of learning (see Fig.~\ref{fig:training_process}). To avoid poor initial decisions and for the smooth transition from an existing scheduler, we adopt offline supervised learning to bootstrap the DRL policy with the existing scheduling strategy. 

\captionsetup[figure]{labelfont=bf}
\begin{figure*}[t]
\centering
\captionsetup{width=0.5\textwidth}
  \includegraphics[width=0.95\textwidth]{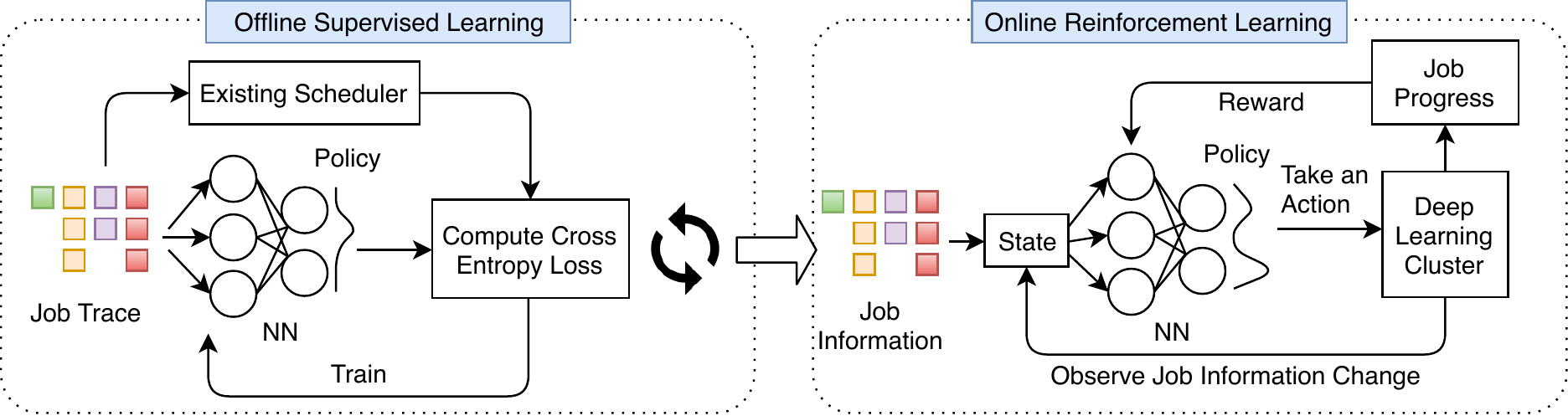}
  \caption{An overview of \DL2}
  \label{workflow}
  \vspace{-3mm}
\end{figure*}

\section{\DL2 Overview}
\label{sec:overview}

The ultimate goal of \DL2~is to find the best resource allocation policy in a live DL cluster and minimize the average job completion time among all concurrent jobs. 

\subsection{DL Cluster} 

In the DL cluster with multiple GPU servers, DL training jobs are submitted over time. Each job runs a distributed ML framework (\eg, MXNet, as in our experiments) to learn a specific DL model by repeatedly training its dataset. 

Upon submission of a job, the user, \ie, job owner, provides her/his resource demand to run each worker and each PS, respectively, as well as the total number of training epochs to run. For example, a worker often requires at least 1 GPU and a PS needs many CPU cores. The total training epoch number to achieve model convergence (\eg, the convergence of loss or accuracy of the model) can be estimated based on expert knowledge or job history. 

Depending on resource availability and training speeds, each job may run over a different number of workers and PSs from one time slot to the other (as decided by the scheduler). For synchronous training, to guarantee the same training result (model) while varying the number of workers, we adjust the mini-batch size of each worker, so that the total batch size in a job, as specified by the user, still remains unchanged~\cite{peng2018optimus}\cite{goyal2017accurate}. For asynchronous training, the mini-batch size of each worker remains the same while the number of workers varies (as the global batch size equals each worker's batch size).

\subsection{\DL2 Scheduler} 

Our DL-based scheduler, \DL2, adopts joint offline and online learning of a policy neural network (NN) for making resource allocation decisions to concurrent jobs in the cluster. An overview of \DL2~is given in Fig.~\ref{workflow}. 

\vspace{1mm}
\noindent {\bf Offline supervised learning.} For warm-up, we use supervised learning to train the policy NN, to initialize a policy whose performance is as good as the existing scheduler in the DL cluster.
A small set of historical job runtime traces collected from the cluster are used for supervised learning, to allow the NN to produce similar decisions as made by the existing scheduler. This step is a must due to the poor performance of applying online RL directly (see Fig.~\ref{fig:training_process}).

\vspace{1mm}
\noindent {\bf Online reinforcement learning.} Online RL works in a time-slotted fashion; each time slot is a scheduling interval, \eg, 1 hour. 
At the beginning of a scheduling interval, the policy NN takes the information of all the concurrent jobs as input state, and produces the numbers of workers and PSs for each job. The concurrent jobs include new jobs arrived in the previous time slot (after previous scheduling) and jobs which were submitted earlier and whose training has not been completed yet. Workers and PSs are placed on physical machines following the placement policy in the cluster, such as load balancing~\cite{randles2010comparative}. 
 Jobs' training progress is observed at the end of each time slot, and used as the reward to improve the policy network. 

\section{Detailed Design} 
\label{sec:joint_learning}

\subsection{Policy Neural Network}

\captionsetup[figure]{labelfont=bf}
\begin{figure}[!t]
\captionsetup{width=0.5\textwidth}
\centering
  \includegraphics[width=3in]{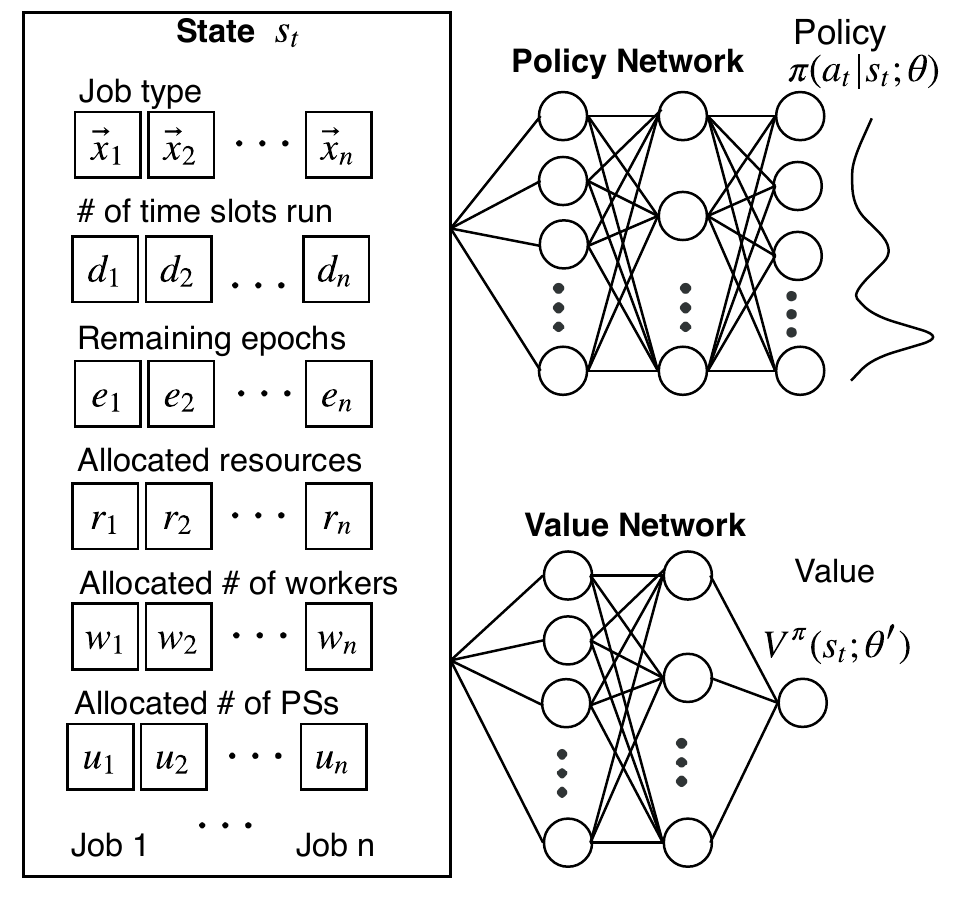}
  \caption{Actor-critic reinforcement learning}
  \label{actor-critic_nn}
  \vspace{-4mm}
\end{figure}


\noindent\textbf{State.} The input state to the policy NN is a matrix $s=(\boldsymbol{x}, \vec{d}, \vec{e}, \vec{r}, \vec{w}, \vec{u})$, including the following (Fig.~\ref{actor-critic_nn}):

$\bullet$ $\boldsymbol{x}$, a $J\times L$ matrix representing the DL models trained in the jobs, where $J$ is an upper bound of the maximal number of concurrent jobs in a time slot that we are scheduling, and $L$ is the maximal number of training job types in the cluster at all times. 
We consider DL jobs training similar DNN architecture as the same type in our input. For example, fine-tuning jobs based on the same pre-trained model is common\footnote{Many computer vision jobs use pre-trained ResNet~\cite{he2016deep} model as initialization for training on a target dataset. Similarly, natural language understanding jobs use BERT~\cite{devlin2018bert} model to initialize training.} and they can be treated as the same type. Each vector $\vec{x}_i$ in $\boldsymbol{x}$, $\forall i=1,\dots,J$, is a one-hot encoding of job $i$'s type. For example, if there are $3$ job types in total and $3$ concurrent jobs in each type respectively, then $\boldsymbol{x}= \{[1 0 0]; [0 1 0]; [0 0 1]\}$.

$\bullet$ $\vec{d}$, a $J$-dimensional vector encoding the number of time slots that each job has run in the cluster, for all jobs. For example, $d_i$ is the number of time slots that job $i$ has run.

$\bullet$ $\vec{e}$, a $J$-dimensional vector encoding the remaining number of epochs to train for each job. $e_i$ is the difference between user-specified total training epoch number for job $i$ and the number of epochs trained till current time slot.

$\bullet$ $\vec{r}$, a $J$-dimensional vector representing the amount of dominant resource already allocated to each job in the current time slot. For example, $r_i$ is the amount of dominant resource (the type of resource the job occupies most as compared to the overall capacity of the resource in the cluster) allocated to job $i$ by resource allocation decisions already made through inferences in this time slot.

$\bullet$ $\vec{w}$ and $\vec{u}$, each of them is a $J$-dimensional vector where the $i$-th item is the number of workers (PSs) allocated to job $i$ in the current time slot. 

Information of concurrent jobs in different components of the state are ordered according to the jobs' arrival times. 
The input state does not directly include available resource capacities in the cluster; our scheduler can handle time-varying overall resource capacities in the cluster. 

\vspace{1mm}
\noindent\textbf{Action.} The NN produces a policy $\pi:\pi(a\mid s;\vec{\theta}) \rightarrow [0,1]$, which is a probability distribution over the action space. $a$ represents an action, and $\vec{\theta}$ is the current set of parameters in the NN. 
A straightforward design is to allow each action to specify the numbers of workers/PSs to allocate to all concurrent jobs; this leads to an exponentially large action space, containing all possible worker/PS number combinations. 
A large action space incurs significant training cost and slow convergence~\cite{vinyals2017starcraft}. 

To expedite learning of the NN, we simplify the action definition, and allow the NN to output an action out of the following $3\times J+1$ actions through each inference: (i) $(i, 0)$, meaning allocating one worker to job $i$, (ii) $(i, 1)$, allocating one PS to job $i$, (iii) $(i, 2)$, allocating one worker and one PS to job $i$, 
(iv) a void action which indicates stopping allocating resources in the current time slot (as allocating more resources does not necessarily lead to higher training speed~\cite{peng2018optimus}). Since each inference only outputs an incremental amount of resources to be allocated to one of $J$ jobs, we allow multiple inferences over the NN for producing the complete set of resource allocation decisions in each time slot: after producing one action, we update state $s$,
and then use the NN to produce another action; the inference repeats until the resources are used up or a void action is produced. The void action indicates that further allocating resources to jobs no longer improves training speeds.

Though we produce the worker/PS numbers for each job anew in each time slot, for a job that has been running in the previous time slot, we compare the new and previous numbers and perform dynamic scaling to adjust the deployment numbers only (\S\ref{sec:implementation}).




\vspace{1mm}
\noindent\textbf{NN architecture.} The input state matrix $s$ 
is connected to a fully connected layer with the ReLU \cite{nair2010rectified} function for activation. The number of neurons in this layer is proportional to the size of the state matrix. 
Output from this layer is aggregated in a hidden fully connected layer, which is then connected to the final output layer. 
The final output layer uses the softmax function \cite{goodfellow2016deep} as the activation function. 
The NN architecture is designed based on empirical training trials. 

 
\subsection{Offline Supervised Learning}
In offline supervised learning, we use stochastic gradient descent (SGD)~\cite{sutton2000policy} to update parameters $\vec{\theta}$ of the policy NN to minimize a loss function, which is the cross entropy of the resource allocation decisions made by the NN and decisions of the existing scheduler in the traces~\cite{mannor2005cross}. The NN is repeatedly trained using the trace data, \eg, hundreds of times as in our experiments, such that the policy produced by the NN converges to the policy of the existing scheduler.

\subsection{Online Reinforcement Learning}

\noindent\textbf{Reward.} \DL2~targets average job completion time minimization in the entire cluster. 
Job completion time would be a natural reward to observe, but it is only known when a job is finished, which may well be hundreds of time slots later. 
The significant feedback delay of the reward is unacceptable for online RL, since the delayed reward provides little guidance to improve the early decisions. 
We 
design a per-timeslot reward to collect more reward samples through the job processes, for more frequent RL model updates to expedite convergence. The per-timeslot reward is the sum of normalized number of epochs that the concurrent jobs have trained in this time slot, where the number of epochs trained in job $i$ ($t_i$) is normalized over the overall number of epochs to train for the job ($E_i$): 

\vspace{-2mm}
{\small
\begin{align}
r_t = \sum_{i\in [J]} \frac{t_i}{E_i}, \forall t=1,\dots,\label{eqn:reward}
\end{align}
}
\vspace{-2mm}
 
\noindent The rationale 
is that the more epochs a job runs in a time slot, the fewer time slots it takes to complete, and hence maximizing cumulative reward amounts to minimizing average job completion time. The normalization is to prevent bias towards large jobs.


\vspace{1mm}
\noindent\textbf{Policy gradient-based learning.} In online RL, the policy NN obtained through offline supervised learning is further trained 
using the REINFORCE algorithm~\cite{williams1992simple}, to maximize the expected cumulative discounted reward $\mathbb{E}[\sum_{t=0}^{\infty}\gamma^t r_t]$, where $\gamma\in(0,1)$ is the discount factor. We model the problem as a non-linear one with long-term impact instead of a traditional linear model with one-round independent feedback, \eg, contextual bandit~\cite{langford2007epoch}, because actions in different time slots are correlated. The REINFORCE algorithm updates the policy network's parameters, $\vec{\theta}$, by performing SGD on $\mathbb{E}[\sum_{t=0}^{\infty}-\gamma^t r_t]$. The gradient is:

\vspace{-2mm}
{\small
\begin{align}
\bigtriangledown_{\vec{\theta}} \mathbb{E}_{\pi}[\sum_{t=0}^{\infty}-\gamma^t r_t]=\mathbb{E}_{\pi}[-\bigtriangledown_{\vec{\theta}}\log(\pi(a\mid s;\vec{\theta}))Q(a,s;\vec{\theta})]\label{eqn:policygradient}
\end{align}
}
\vspace{-2mm}

\noindent where the Q value, $Q(a,s;\vec{\theta})$ represents the ``quality'' of the action $a$ taken in given state $s$ following the policy $\pi(\cdot;\vec{\theta})$, calculated as the expected cumulative discounted reward to obtain after selecting action $a$ at state $s$ following $\pi(\cdot;\vec{\theta})$. 
Each Q value can be computed (empirically) using a mini-batch of samples~\cite{sutton2000policy}. Each sample is a four-tuple, $(s,a,s',r)$, where $s'$ is the new state after action $a$ is taken in state $s$. 

Note that our system runs differently from standard RL: we do multiple inferences (\ie, produce multiple actions) using the NN in each time slot $t$; the input state changes after each inference; we only observe the reward and update the NN once after all inferences in the time slot are done. We can obtain multiple samples in a time slot $t$, and set the reward in each sample to be the reward (\ref{eqn:reward}) observed after all inferences are done in $t$.


\vspace{1mm}
We further adopt a number of techniques to stabilize online RL, expedite policy convergence, and improve the quality of the obtained policy.

\vspace{1mm}
\noindent\textbf{Actor-critic.} 
We improve the basic policy gradient-based RL with the actor-critic algorithm~\cite{mnih2016asynchronous} (illustrated in Fig.~\ref{actor-critic_nn}), 
 for faster convergence of the policy network. The basic idea is to replace Q value in Eqn.~\ref{eqn:policygradient} with an advantage, $Q(a,s;\vec{\theta})-V^{\pi}(s,\vec{\theta})$, where $V^{\pi}(s,\vec{\theta})$ is a value function, representing the expected reward over the actions drawn using policy $\pi(a\mid s;\vec{\theta})$ at all times starting from time slot $t$. The advantage shows how much better a specific action is, as compared to the expected reward of taking actions according to $\pi(a\mid s;\vec{\theta})$ in the current state. Using the advantage in computing the policy gradients ensures a much lower variance in the gradients, such that policy learning is more stable.


The value function is evaluated by a value network, which has the same NN structure as the policy network except that its final output layer is a linear neuron without any activation function \cite{mnih2016asynchronous}, and it produces the estimate of value function $V^{\pi}(s,\vec{\theta})$. The input state to the value network is the same as that to the policy network. 
We train the value network using the temporal difference method~\cite{mnih2016asynchronous}.


\vspace{1mm}
\noindent\textbf{Job-aware exploration.}
To obtain a good policy through RL, we need to ensure that the action space is adequately explored (\ie, actions leading to good rewards can be sufficiently produced); as otherwise, the RL may well converge to poor local optimal policy~\cite{vinyals2017starcraft}\cite{mnih2016asynchronous}. We first adopt a commonly used entropy exploration method, by adding an entropy regularization term $\beta \bigtriangledown_{\vec{\theta}}H(\pi(\cdot \mid s ; \vec{\theta}))$ in gradient calculation to update the policy network~\cite{mnih2016asynchronous}. 
In this way, parameters of the policy network, $\vec{\theta}$, is updated towards the direction of higher entropy (implying exploring more of the action space).

During training, we observe a large number of unnecessary or poor explorations (\eg, allocating multiple workers but 0 PS to a job) due to unawareness of job semantics. To 
improve exploration efficiency, we adopt another technique based on the $\epsilon$-greedy method~\cite{sutton1998reinforcement}. At each inference using the policy network, we check the input state: if the input state belongs to one of the poor states that we have identified, with probability $1-\epsilon$, we apply the resource allocation decisions produced by the policy network, and with probability $\epsilon$, we discard the output from the policy network, but adopt a specified action and observe the reward of this action. 

The set of poor input states includes three cases: (i) there exists one job to be scheduled which has been allocated with multiple workers but no PS; (ii) there exists one job which has been allocated multiple PSs but no workers; (iii) there exists one job whose allocated numbers of workers ($w$) and PSs ($u$) differ too much, \ie, $w/u>threshold$ or $u/w>threshold$ (the threshold is 10 in our experiments). Our manually specified action upon each of these input states is: (i) 
allocate one PS to that job; (ii) allocate one more worker to the job; (iii) allocate one more PS or one more worker to that job, to make its worker/PS numbers more even. 

\vspace{1mm}
\noindent\textbf{Experience replay.}
It is known that correlation among the samples prevents convergence of an actor-critic model to a good policy~\cite{sutton1998reinforcement}. 
In our online RL, the current policy network determines the following training samples, \eg, if the policy network finds that allocating more workers improves reward, then the following sample sequence will be dominated by those produced from this strategy; this may lead to a bad feedback loop, preventing the exploration of samples with higher rewards. 

To alleviate correlation in the observed sample sequence, we adopt experience replay~\cite{mnih2015human} in the actor-critic framework. Specifically, we maintain a replay buffer to store the samples collected in the latest time slots. At the end of each time slot, instead of using all samples collected during this time slot, we select a mini-batch of samples from the replay buffer to compute the gradient updates, where the samples could be from multiple previous time slots. 
\section{Dynamic Scaling}
\label{sec:implementation}

Though node addition and deletion are supported in system design in the literature~\cite{li2014scaling}\cite{harlap2017proteus}\cite{qiao2018litz}
, existing open-source distributed machine learning frameworks (\eg, TensorFlow~\cite{abadi2016tensorflow}, MXNet~\cite{chen2016mxnet}, Caffe~\cite{Caffe2}
) do not support dynamic worker/PS adjustment in a running job. To adjust the number of workers/PSs in a job, a simple and general approach is checkpointing (\eg, Optimus~\cite{peng2018optimus}): terminate a training job and save global model parameters as a checkpoint image; 
 then restart the job with a new deployment of PSs and workers, 
 and the saved model parameters. Checkpointing and restarting add additional delay to the training process~\cite{qiao2018litz}. For example, it takes 1 minute to checkpoint and stop training, and another 5 minutes to completely restore training of a DSSM model~\cite{shen2014latent}, due to data re-preprocessing before training starts. The overhead is significant when the frequency of resource scaling is high (\eg, every hour). The other approach is to resize resources without terminating training process. As an example, we improve the MXNet framework \cite{chen2016mxnet} to enable dynamic ``hot'' scaling.

\vspace{1mm}
\noindent\textbf{Challenges.} In the parameter server architecture, each PS maintains a subset of the parameters in the global model. When the number of PSs changes, the global parameters need to be migrated among the PSs (for load balancing), and workers should be informed in time to send parameter updates to the correct PSs. When the number of workers changes, the new connections between new workers and the PSs should be established. The key challenges are: (1) {\textit{correctness}}, \ie, a consistent copy of the global model parameters should be maintained while parameters are moved across the PSs, and workers always send gradients to correct PSs; (2) {\textit{high performance}, \ie, we should ensure that interruption to training is minimized and the PSs are load balanced.


\vspace{1mm}
\noindent\textbf{Scaling Steps.} We add a {\em coordinator} module into the MXNet framework, which works with \DL2~scheduler to handle joining of new workers or PSs and termination of existing ones. We demonstrate our design using the case of adding a new PS into an existing job. The steps are shown in Fig.~\ref{fig:ps_scaling}. 

\begin{figure}[t]        
\center{\includegraphics[width=\linewidth]  {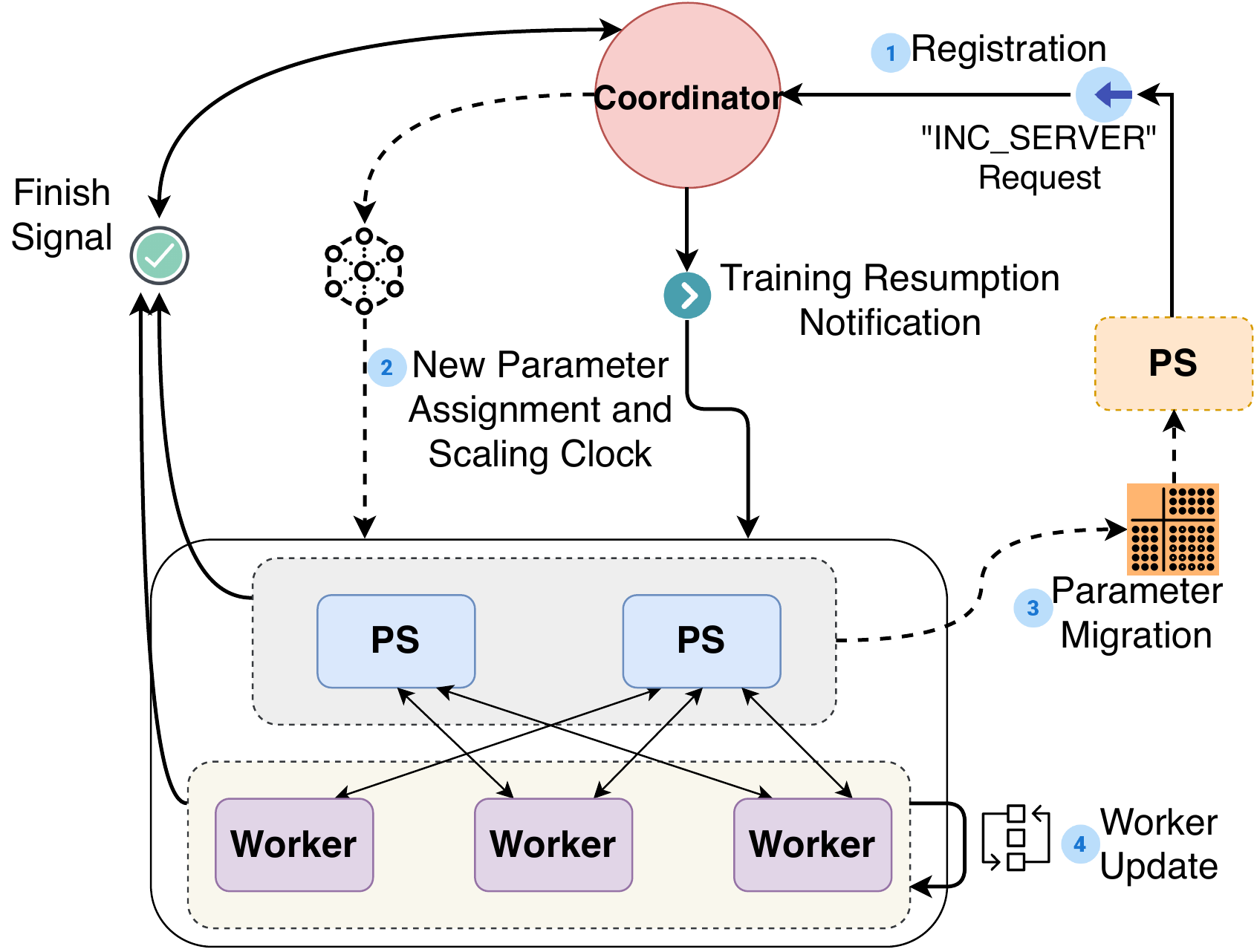}}        
\caption{Steps for adding one PS into a running job}
\label{fig:ps_scaling} 
\vspace{-5mm}
\end{figure}


\vspace{1mm}
{\noindent{\em 1) Registration.}}
When a new PS is launched, it registers itself with the coordinator by sending an ``INC\_SERVER'' request message. The PS will then receive its ID in the job, the global parameters it is responsible to maintain, and the current list of workers and PSs to establish connections with. After that, the PS starts functioning, awaiting workers' parameter updates and further instructions from the coordinator (\eg, parameter migration).

\vspace{1mm}
{\noindent{\em 2) Parameter assignment.}}
Upon receiving a registration request, the coordinator 
 updates its list of workers and PSs, and computes parameter assignment to the new PS. A best-fit algorithm is adopted: 
 move part of the parameters on each existing PS to the new PS, such that all PSs maintain nearly the same number of parameters, while 
 minimizing parameter movement across the PSs. 

In order to keep a consistent copy of global model parameters when migrating parameters among PSs, we maintain a version counter for parameters. For PSs, the version counter is the number of parameter updates; for workers, the version counter is received from PSs when pulling updated parameters. To decide when PSs should migrate parameters, we calculate a scaling clock based on current version counter and round trip time between the coordinator and PSs/workers.



The coordinator sends new parameter assignment among PSs and the scaling clock to all PSs and workers.%


\vspace{1mm}
{\noindent{\em 3) Parameter migration.}}
 At each PS, when the version counter of parameters reaches the scaling clock received from the coordinator, 
the PS moves its parameters to the new PS according to the parameter assignment decisions received\footnote{For asynchronous training, the PS may need to buffer push or pull requests from workers and redirect them to the new PS.}. 
Once parameter migration among all PSs is completed, the coordinator notifies all workers to resume training.

\vspace{1mm}
{\noindent{\em 4) Worker update.}} 
At each worker, once its version counter equals the scaling clock received from the coordinator, the worker suspends its push/pull operations and awaits notification for completion of parameter migration. Upon notification from the coordinator, the workers update their parameter-PS mapping, establish connections with the new PS, and resume the training process. 

\vspace{1mm}
In case of removing a PS, the scheduler chooses the PS to be removed by keeping the load balanced among the physical machines.
The chosen PS sends a removal request to the coordinator. Similar steps as 2)3)4) above are then carried out, except that parameters in the removed PS are moved to other PSs, using the best-fit algorithm.

To add a new worker into an existing job, the coordinator 
sends the current parameter-PS mapping in the response to the worker's registration message. It also notifies all PSs the addition of the new worker for building connections. 
The worker starts operation after training dataset is copied. For worker removal, the scheduler chooses the worker to be removed by keeping the load balanced across physical machines.
The coordinator receives a removal request from the worker, and then broadcasts it to all workers and PSs for updating their node lists. The mini-batch size of workers is adjusted so as to keep total batch size the same.

\section{Evaluation}
\label{sec:evaluation}

\subsection{\DL2 Implementation}
We implement \DL2~as a custom scheduler on Kubernetes~\cite{kubernetes}. We run each training job using the MXNet framework~\cite{chen2016mxnet}. Workers and PSs are running on Docker containers. Training data of jobs are stored in HDFS 2.8~\cite{hdfs}. The scheduler constantly queries cluster resources and job states (\eg, training speeds) and instructs deployment of a new job or resource adjustment in an existing job via Kubernetes API server. Mapping the cluster and job states to a scheduling decision takes less than 3ms.


For each new job, \DL2~launches its coordinator, workers, and PSs on machines decided by the default placement strategy of the cluster (\ie, load balancing). The coordinator is informed of the workers and PSs in the job via Kubernetes API. When a worker/PS container is launched on a machine, an agent in the container starts execution. It queries the readiness of other containers of the same job via Kubernetes API and starts user-provided training scripts after all other containers are ready. The agent also monitors the training status, \eg, the number of trained steps, accuracy, and training speed.

\subsection{Methodology}

{\noindent\textbf{Testbed.}}
Our testbed includes 13 GPU/CPU servers connected by a Dell Networking Z9100-ON 100GbE switch. Each server has one Intel E5-1660 v4 CPU, two GTX 1080Ti GPUs, 48GB RAM, one MCX413A-GCAT 50GbE NIC, one 480GB SSD, and one 4TB HDD. Each server runs Ubuntu 14.04 LTS and Docker 17.09-ce~\cite{docker}. 

\begin{table}[t]
\centering
\small
\caption{DL Jobs in Evaluation}
\label{table:job_details}
\begin{tabular}{|c|c|c|}
\hline
Model										& Application domain 		& Dataset	\\
\hline
ResNet-50~\cite{he2016deep}					& image classification		& ImageNet~\cite{imagenet}	\\ 	
VGG-16~\cite{simonyan2014very}								& image classification		& ImageNet~\cite{imagenet}	\\ 	
ResNeXt-110~\cite{he2017resnext}			& image classification		& CIFAR10~\cite{cifar10_dataset}		\\	
Inception-BN~\cite{szegedy2016rethinking}	& image classification		& Caltech~\cite{caltech_dataset}		\\ 
Seq2Seq~\cite{gehring2017convolutional}		& machine translation		& WMT17~\cite{wmt17_dataset}			\\	
CTC~\cite{kim2014convolutional}		& sentence classification	& mr~\cite{mr_dataset}					\\	
DSSM~\cite{shen2014latent}					& word representation		& text8~\cite{text8_dataset}			\\	
WLM~\cite{word_language_model}	& language modeling			& PTB~\cite{ptb_dataset}				\\	
\hline
\end{tabular}
\vspace{-1mm}
\end{table}

{\noindent\textbf{Trace.}}
We use patterns from a 75-day real-world job trace collected from a large production DL cluster with a few thousands of GPUs and thousands of jobs, to drive our testbed experiments and simulation studies. 
 Fig.~\ref{fig:trace_sketch}~(a) shows the job arrival rate (number of jobs arrived per time slot, \ie, 20 minutes) during a typical week. 
 Fig.~\ref{fig:trace_sketch}~(b) shows the distribution of job duration: over a half of jobs run for more than an hour and some for days; the average job duration is 147 minutes. 
 
 Due to security and privacy concerns of the company, the job source code is not available, and we do not know job details (\eg, model architecture). So we select 8 categories of ML models for experiments, from official MXNet tutorials~\cite{mxnet_official_examples}, 
 with representative application domains, different architectures and parameter sizes \cite{mxnet_official_examples}, as shown in Table~\ref{table:job_details}. Each worker in different jobs uses at most 2 GPUs and 1-4 CPU cores, and each PS uses 1-4 CPU cores.

In both testbed experiments and simulations, the jobs are submitted to the cluster following the dynamic pattern in Fig.~\ref{fig:trace_sketch}~(a) (with arrival rates scaled down). Upon an arrival event, we randomly select a model from Table~\ref{table:job_details} and vary its required number of training epochs (tens to hundreds) to generate a job variant, following job running time distribution of the real-world trace (scaled down). For models training on large datasets (\eg, ImageNet~\cite{imagenet}), we downscale the datasets so that the training can be finished in a reasonable amount of time. In experiments, 30 jobs are submitted to run in our testbed;
 in simulations, 500 servers are simulated, and 200 jobs are submitted in the simulated cluster.

{\noindent\textbf{Training setting.}} Our DL-based scheduler is implemented using TensorFlow~\cite{abadi2016tensorflow}. The neural network is trained using Adam optimizer~\cite{kingma2014adam} with a fixed learning rate of $0.005$ for offline supervised learning and $0.0001$ for online reinforcement learning, mini-batch size of $256$ samples, reward discount factor $\gamma=0.9$, exploration constant $\epsilon=0.4$, entropy weight $\beta=0.1$, 
and an experience replay buffer of $8192$ samples. The network has $2$ hidden layers with $256$ neurons each. These hyper-parameters (neural network structure, learning rate, mini-batch size, etc.) are chosen based on a few empirical training trials. We refer to one update of the neural network at the end of each time slot as one {\em step} in this section.
 
{\noindent\textbf{Baselines.}}
We compare \DL2~with the following baselines.

$\bullet$ Dominant Resource Fairness (DRF)~\cite{ghodsi2011dominant}: It allocates resources to jobs based on the fairness of dominant resources. By default, we use DRF as the existing scheduler used to guide supervised learning in \DL2, since it is widely adopted in existing cluster schedulers, \eg, YARN~\cite{vavilapalli2013apache}, Mesos~\cite{hindman2011mesos}.

$\bullet$ Tetris~\cite{grandl2014multi}: It preferentially allocates resources to jobs with the shortest remaining completion time and highest resource packing efficiency.

$\bullet$ Optimus~\cite{peng2018optimus}: It is a customized scheduler for DL workloads, which builds a performance model for deep learning jobs to estimate remaining training time and adopts a greedy heuristic to schedule jobs. 

$\bullet$ OfflineRL: The offline reinforcement learning algorithm adopts pure offline training, under the same training settings as online RL in \DL2, except that the training data are generated by an analytical performance model~\cite{peng2018optimus} in a simulation environment (we do not use the trace as it does not contain feedback to all decisions the offline training produces).


Wherever appropriate, we use separate training dataset and validation dataset. Both include job sequences generated using the job arrival and duration distributions from the trace. The random seeds are different when generating the datasets, to ensure that they are different. 


\begin{figure}
\centering
\begin{minipage}[hb]{0.495\linewidth}
\centering
\subcaptionbox{Job arrival rate in a week}
{\includegraphics[width=\linewidth]{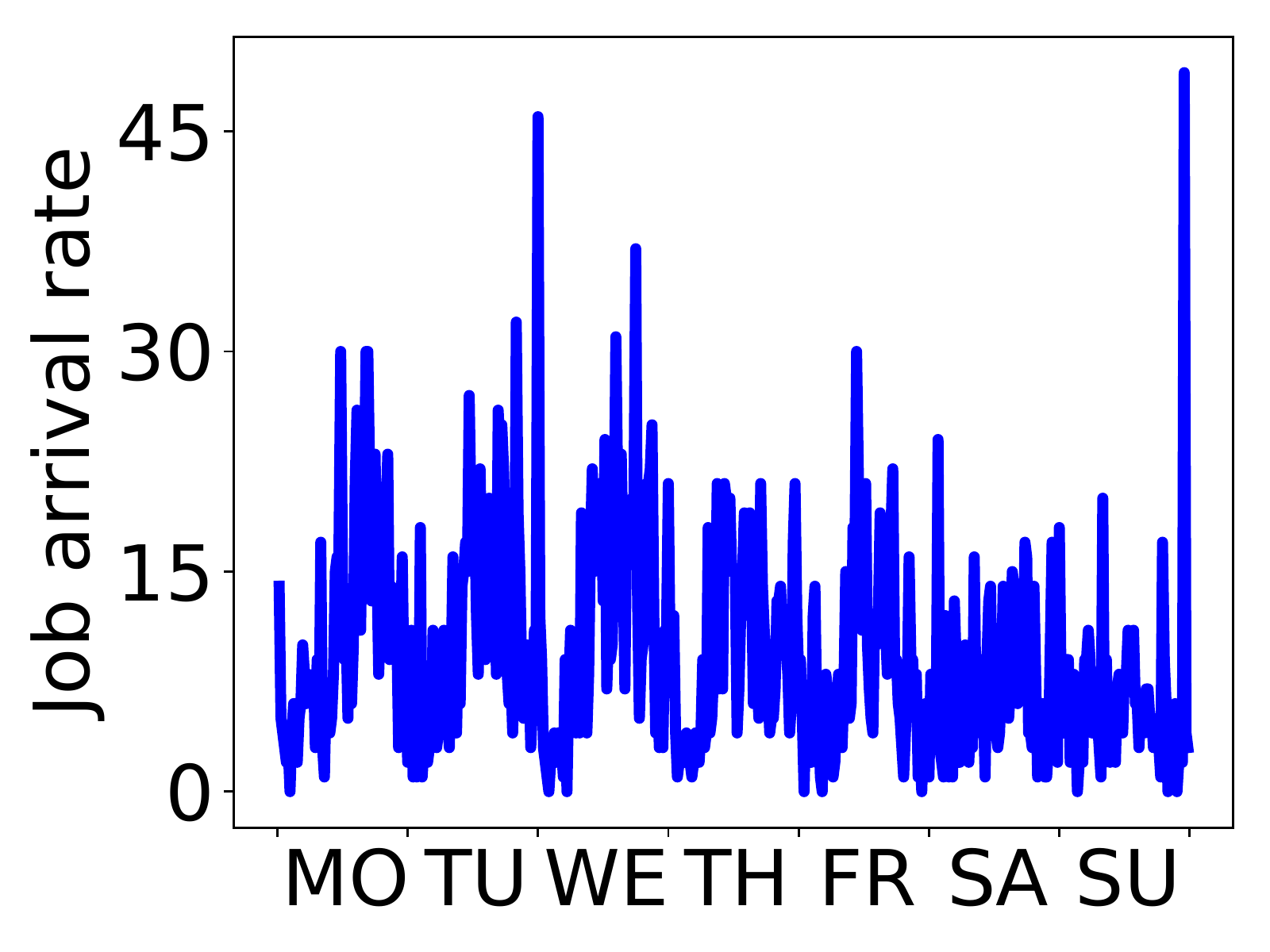}}
\end{minipage}%
\begin{minipage}[hb]{0.495\linewidth}
\centering
\subcaptionbox{CDF of job duration}
{\includegraphics[width=\linewidth]{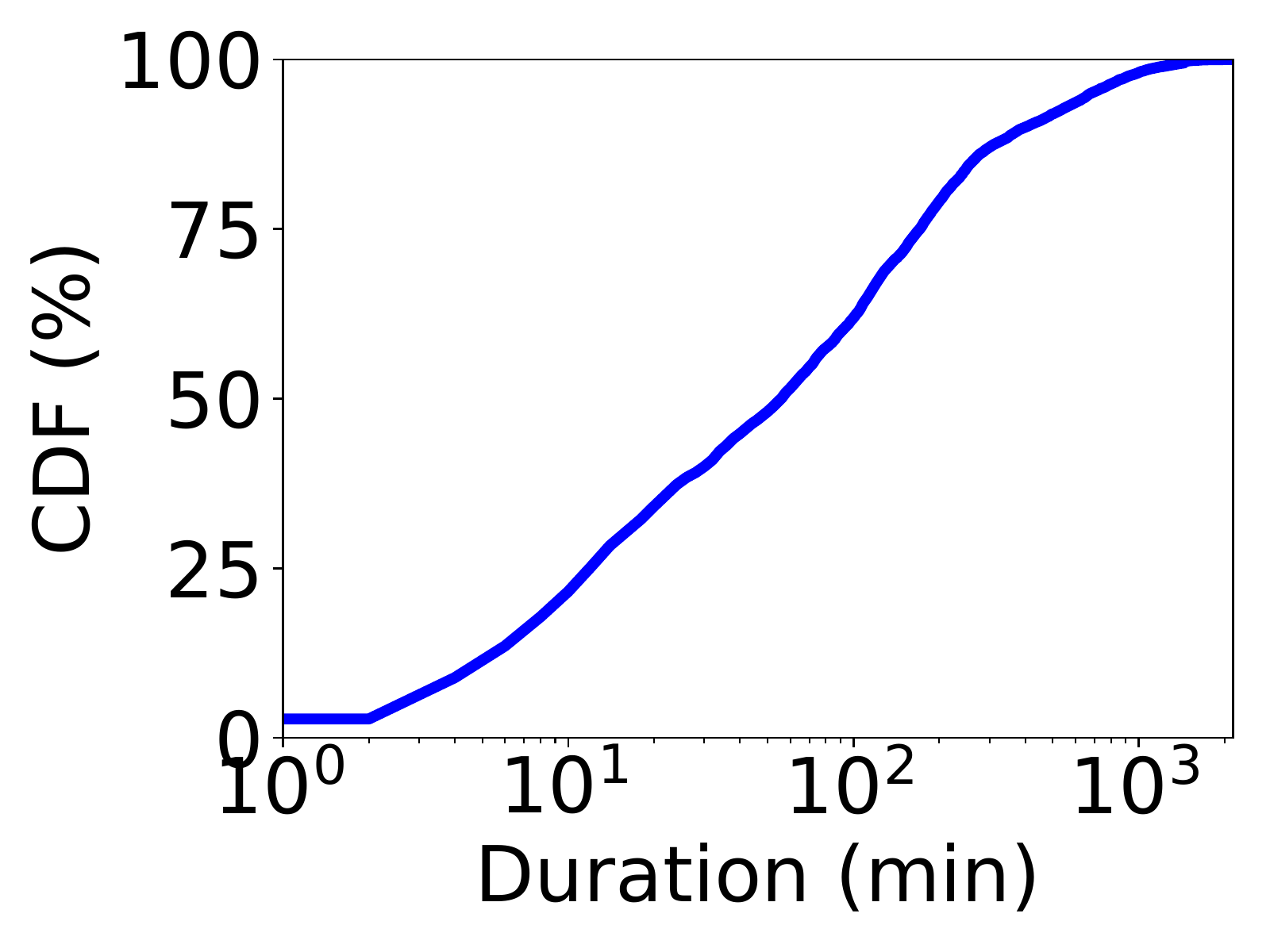}}
\end{minipage}
\caption{Trace sketch}
\label{fig:trace_sketch}
\vspace{-4mm}
\end{figure}

\subsection{Performance}

We first compare the performance of \DL2~with baselines and show the overhead of dynamic scaling using testbed experiments. 

{\noindent\textbf{Comparison.}} 
Fig.~\ref{fig:performance_comparison} shows that \DL2~improves average job completion time by $44.1\%$ 
 when compared to DRF. 
 Tetris performs better than DRF but worse than \DL2: once it selects a job with the highest score in terms of resource packing and remaining completion time, it always adds tasks to the job until the number of tasks reaches a user-defined threshold. 
 When compared to Optimus, \DL2~achieves $17.5\%$ higher performance, since Optimus' estimation of training speed is inaccurate due to cluster interference and evolved MXNet framework (\eg, communication does not overlap with backward computation in Optimus' model). \DL2~also outperforms OfflineRL by $37.9\%$ due to its online training using realistic feedback.

For a better understanding of \DL2's performance gain, Fig.~\ref{fig:training_process} shows how the validated performance keeps improving during the training process, when the policy NN is trained using offline supervised learning only (green curve), online RL only (cyan curve), and offline supervised learning followed by online RL (green+blue). The average job completion time shown at each time slot (\ie, step) is computed over job sequence in the validation dataset, using the policy network trained (on the training dataset) at the current step. 
 We see that with pure online RL, it takes hundreds of steps to achieve the same performance of DRF; with offline supervised learning, the performance quickly converges to a point that is close to DRF's performance within tens of steps (\ie, model updates); as we continue training the NN using online RL, the performance further improves a lot. The performance of DRF is fixed as its strategy does not change over time. Besides smaller job completion time, we also observe that \DL2~has higher CPU and GPU utilization (similar observation as in~\cite{peng2018optimus}). 


\begin{figure}[t]
\centering
\begin{minipage}[t]{0.495\linewidth}
\centering
\includegraphics[height=1.2in]{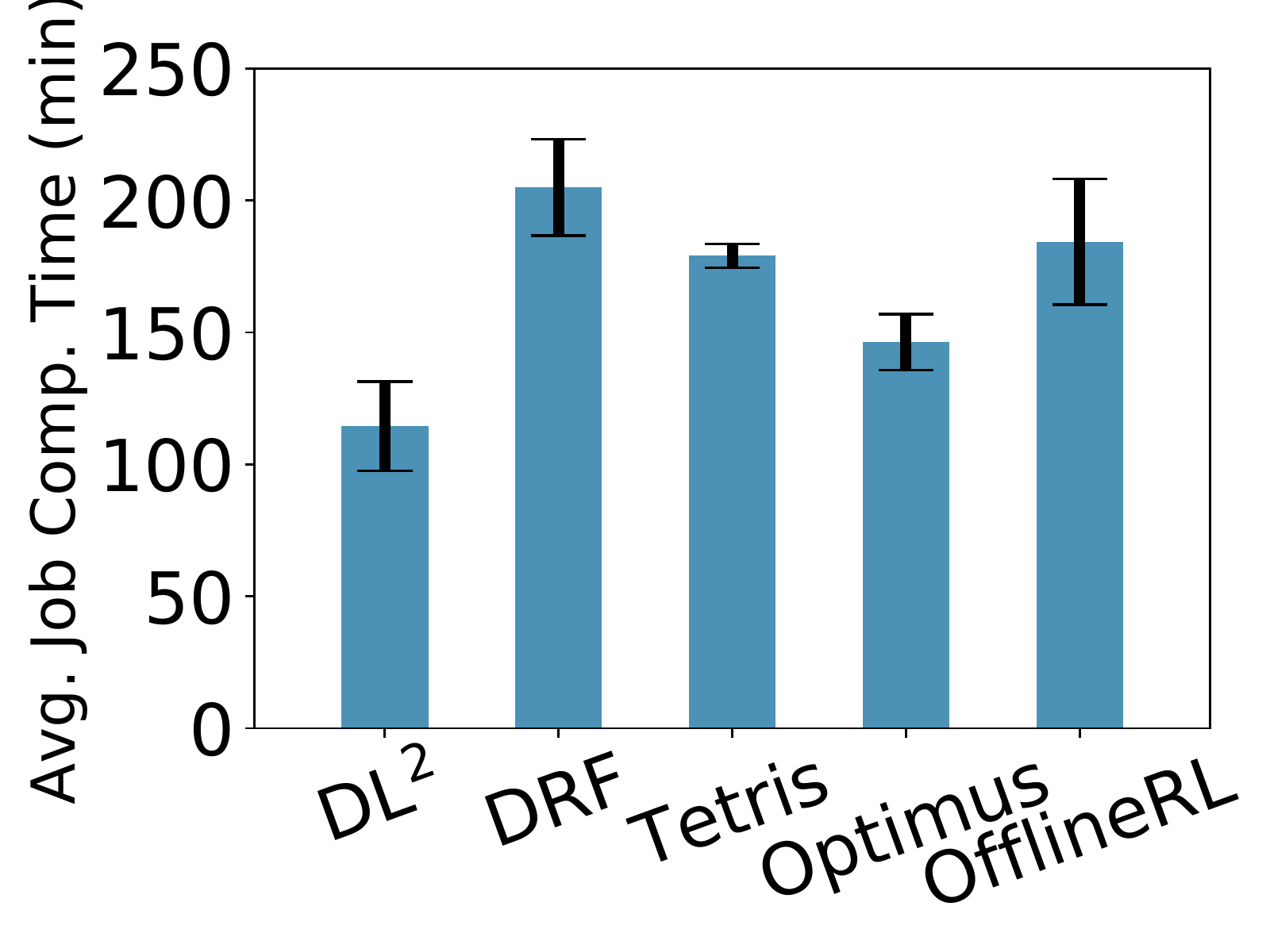}
\captionof{figure}{Performance\\ comparison} 
\label{fig:performance_comparison}
\end{minipage}%
\begin{minipage}[t]{0.495\linewidth}
\centering
\includegraphics[height=1.2in]{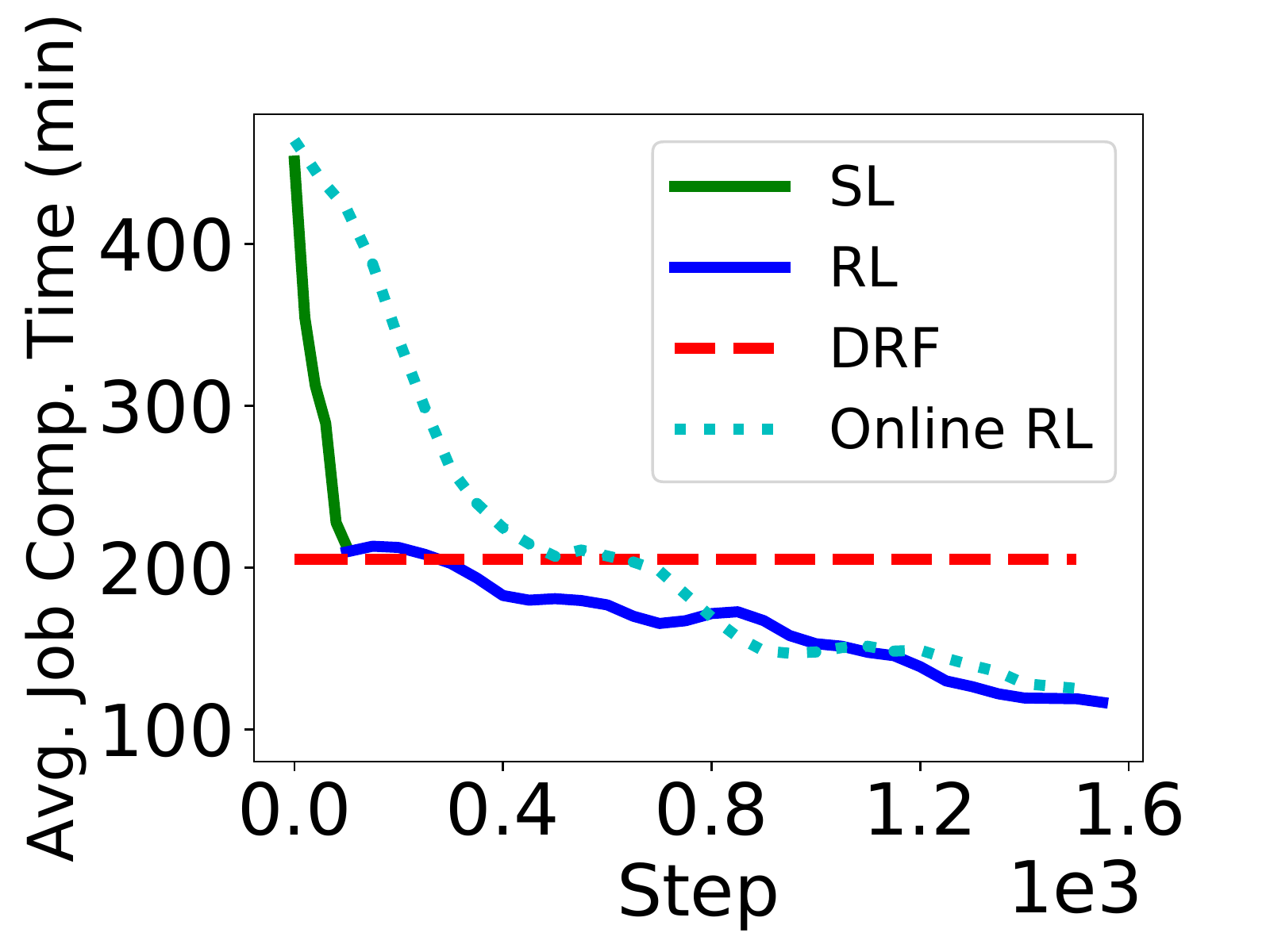}
\captionof{figure}{Training \\progress comparison}
\label{fig:training_process}
\end{minipage}%
\vspace{-4mm}
\end{figure}

{\noindent\textbf{Scaling overhead.}}
Fig.~\ref{fig:scaling_overhead_comparison} compares the average training suspension time among all workers when checkpointing and our scaling approach are used, when different numbers of PSs are added to a ResNet-50~\cite{he2016deep} job. 
The training suspension duration at a worker in \DL2~is from when all the received iteration counts from PSs becomes equal to the scaling clock the worker received from the coordinator, to when the worker resumes training. 
The checkpoint-based approach takes tens of seconds due to model saving, container relaunching and initialization before restarting training. 
The overhead in \DL2~is very small (\ie, tens of milliseconds), even if the time increases linearly with the number of PSs (since we add PSs one by one). We observe similar overhead when removing PSs. On the other hand, adding or removing workers brings little interruption to existing workers, which continue with their training until adjusted training datasets are copied. 

We examine detailed time cost for the 4 steps during the scaling process (\S\ref{sec:implementation}) for adding a PS when training different models. In Fig.~\ref{fig:scaling_step_timing}, the models are listed in increasing order of their model sizes. We observe that the scaling process spends most time in step 3 and step 4, while the time for step 1 and step 2 is negligible. The larger a model is, the more time is spent on parameter movement (step 3). Note that only step 4 blocks worker training and is considered as overhead when compared to checkpointing. Step 3 and step 4 may happen concurrently. 

\begin{figure}[t]
\centering
\begin{minipage}[t]{\linewidth}
\centering
\includegraphics[width=\linewidth]{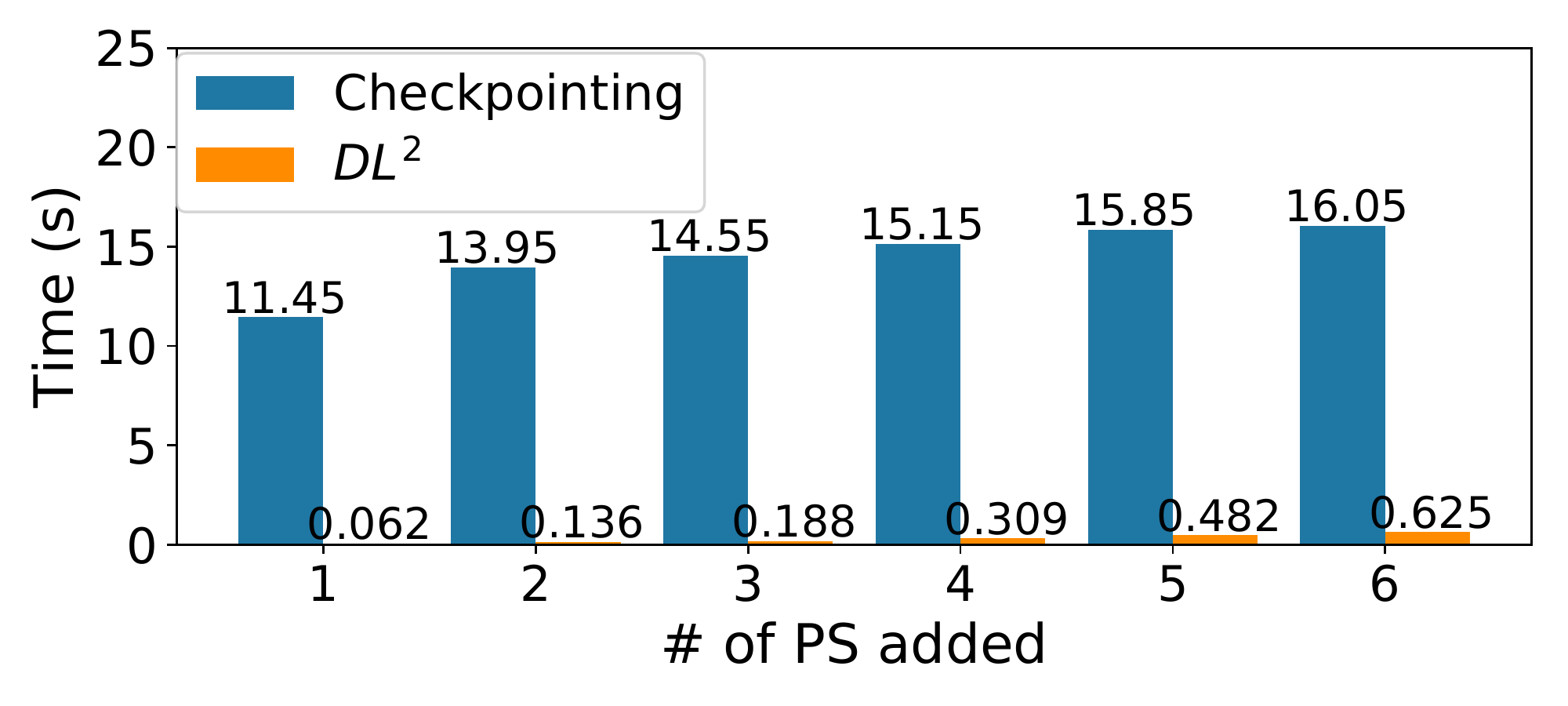}
\captionof{figure}{Scaling overhead comparison} 
\label{fig:scaling_overhead_comparison}
\end{minipage}%
\vspace{-4mm}
\end{figure}
\begin{figure}[t]
\centering
\begin{minipage}[t]{\linewidth}
\centering
\includegraphics[width=\linewidth]{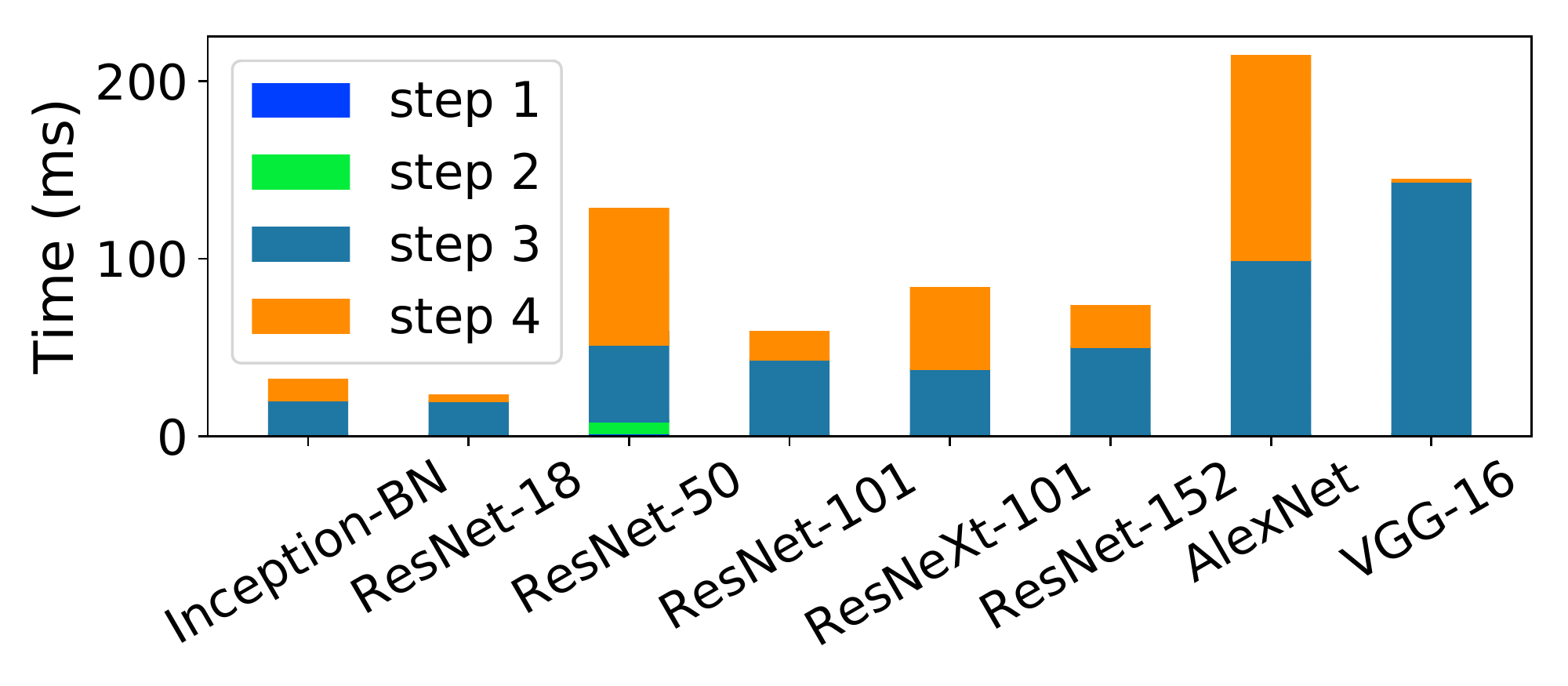} 
\captionof{figure}{Timing of scaling steps} 
\label{fig:scaling_step_timing}
\end{minipage}%
\vspace{-4mm}
\end{figure}

In the following, we carry out controlled large-scale simulations to examine various aspects of \DL2 design. 

\subsection{Generality}
{\noindent\textbf{Training completion time variation.}} To see how \DL2~handles practical performance variation (which white-box schedulers may not handle well), we vary the training speeds in each type of jobs to simulate variation in the training completion time of the same type of jobs (the total numbers of epochs to train remain the same). In Fig.~\ref{fig:jct_vary_train_speed_error}, the variation indicates how the training speed deviates from the average speed (which can be faster or slower by the respective percentage). 
 We see that Optimus is more sensitive to the variation, 
 as it can be easily stuck in local optimum: its scheduling relies on the convexity of the performance model, but training speed variation often breaks convexity. The average job completion time shown in all simulation figures is in time slots.

{\noindent\textbf{Total training epoch estimation.}} \DL2~uses the total number of training epochs of jobs as input, estimated by users. 
 The estimated total number of epochs may well be different from the actual numbers of epochs the jobs need to train to achieve model convergence. 
 We examine how \DL2~performs under different estimation errors: suppose $v$ epochs is fed into \DL2~as the total epoch number that a job is to train, but $v\cdot(1+error)$ or $v\cdot(1-error)$ is the actual number of trained epochs for the job's training convergence. Fig.~\ref{fig:jct_vary_epoch_est_error} shows that the average job completion time increases slightly when the estimation error is larger. It still outperforms DRF (which is oblivious of the estimation errors) by $28\%$ when the error is $20\%$.

\begin{figure}[t]
\centering
\begin{minipage}[t]{0.495\linewidth}
\centering
\includegraphics[height=1.2in]{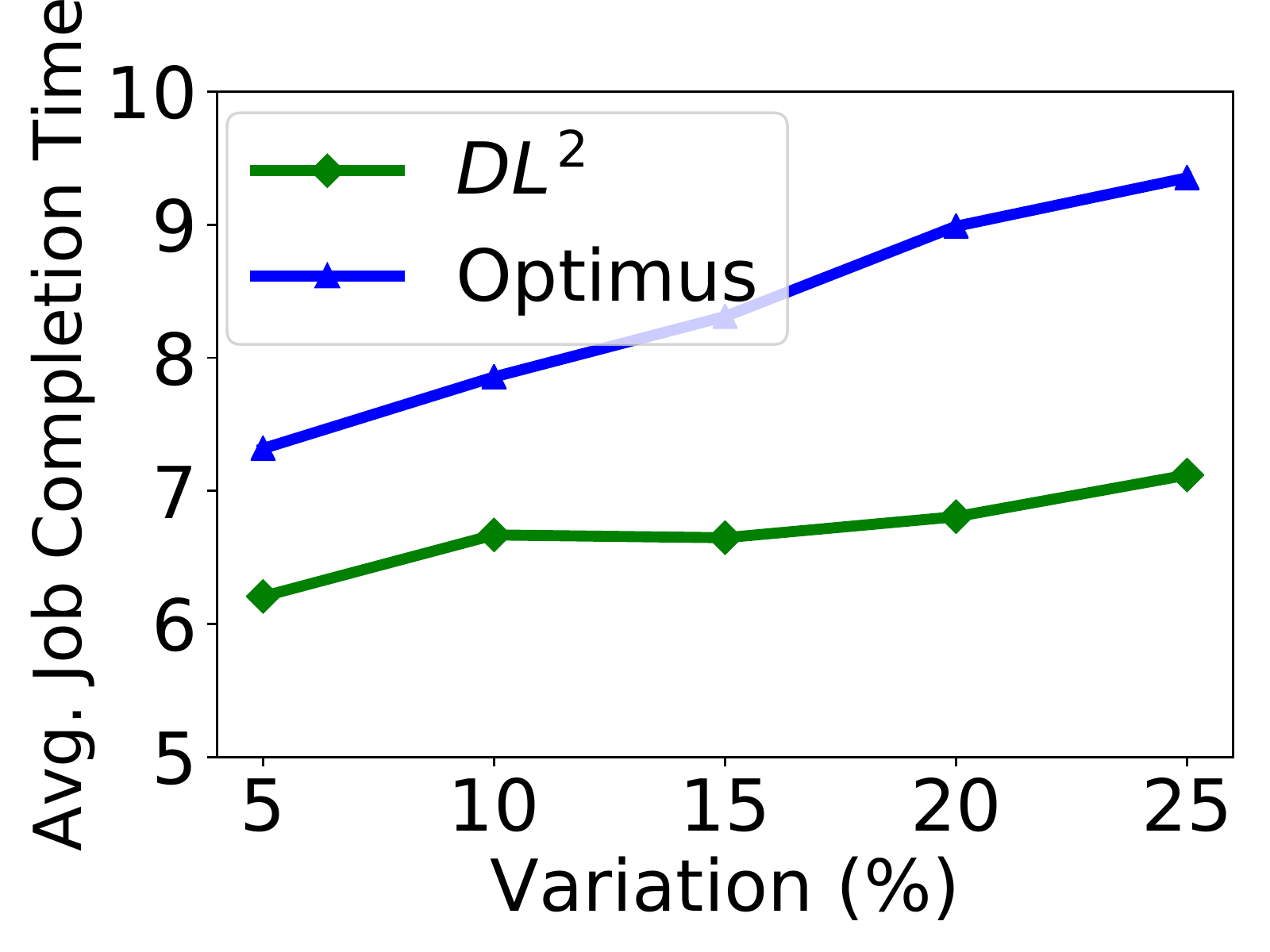}
\captionof{figure}{Training \\completion time variation} 
\label{fig:jct_vary_train_speed_error}
\end{minipage}%
\begin{minipage}[t]{0.495\linewidth}
\centering
\includegraphics[height=1.2in]{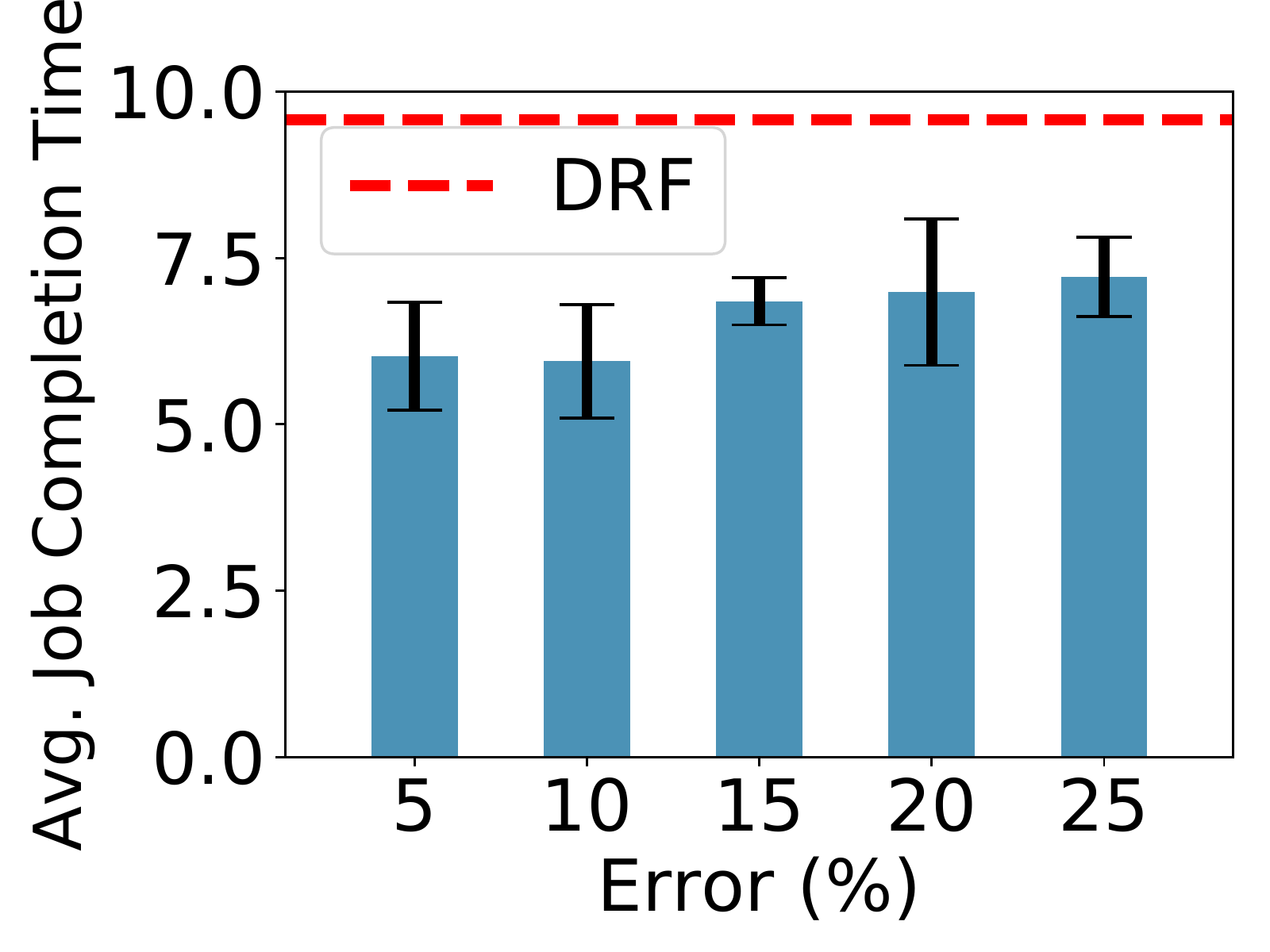}
\captionof{figure}{Performance under diff. epoch estimation errors}
\label{fig:jct_vary_epoch_est_error}
\end{minipage}%
\vspace{-4mm}
\end{figure}

\begin{figure}[t]
\centering
\begin{minipage}[t]{0.495\linewidth}
\centering
\includegraphics[height=1.2in]{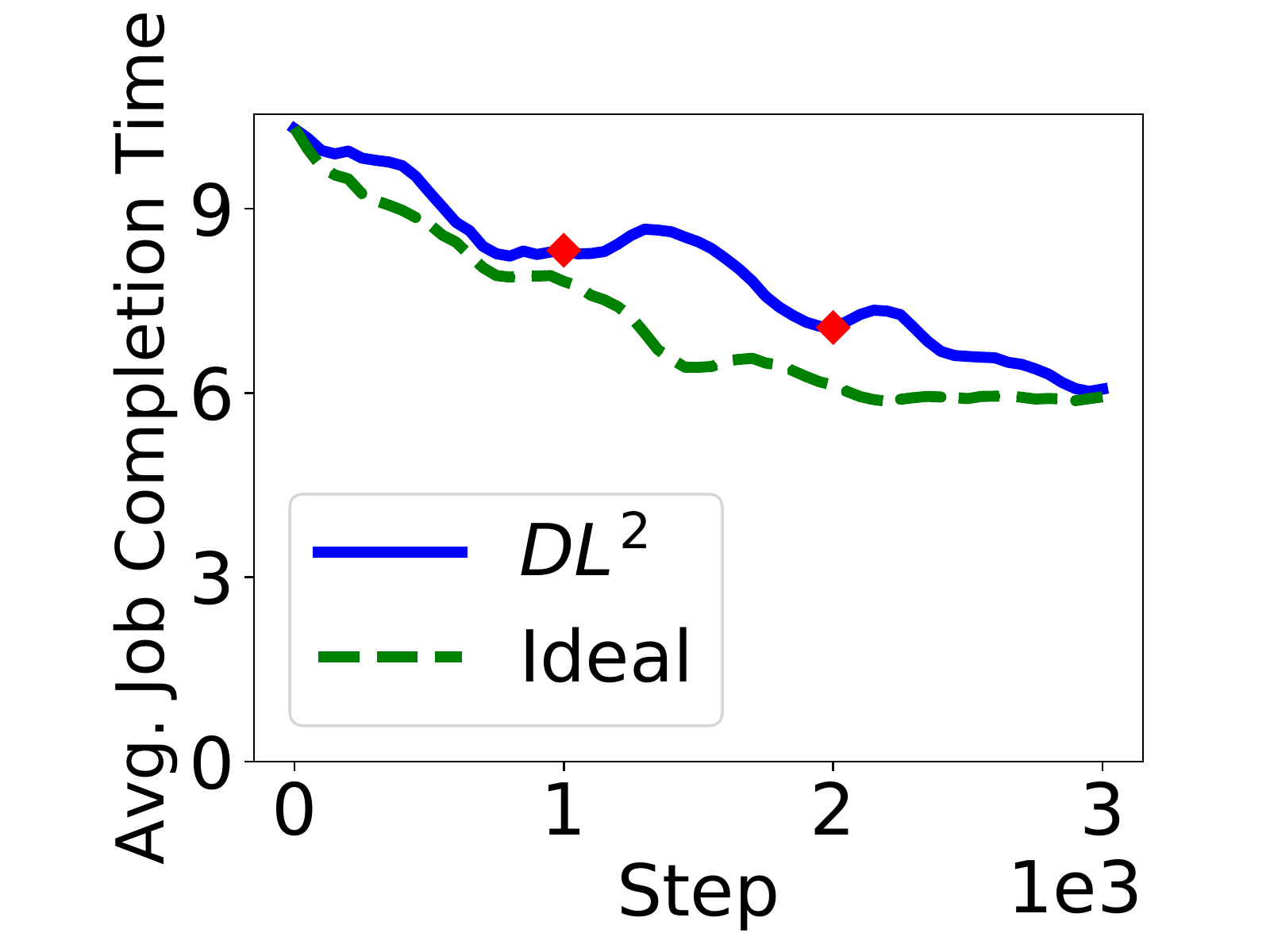}
\captionof{figure}{Handling \\new types of jobs}
\label{fig:unseen_models}
\end{minipage}%
\begin{minipage}[t]{0.495\linewidth}
\centering
\includegraphics[height=1.2in]{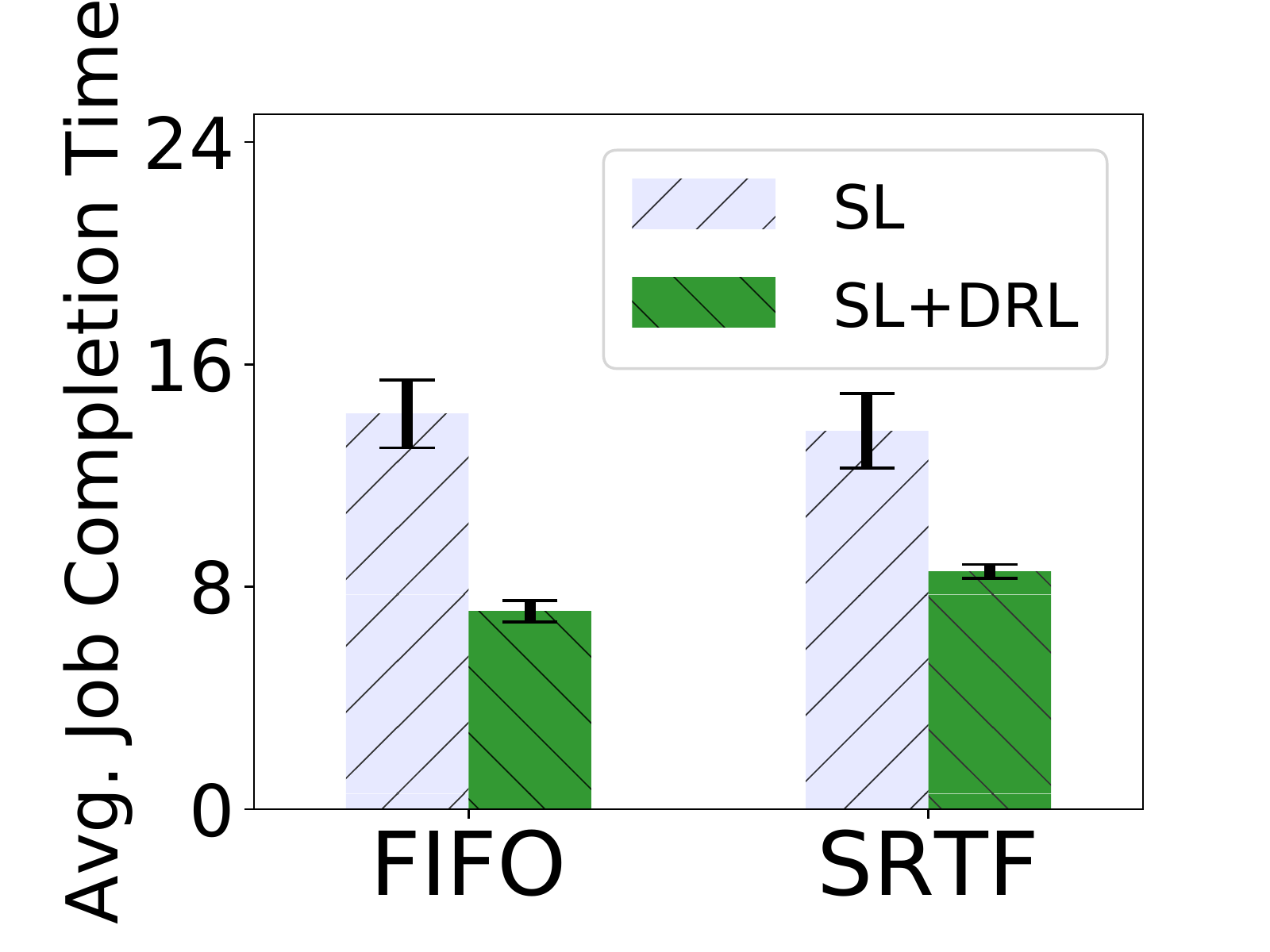}
\captionof{figure}{Different existing scheduling strategies} 
\label{fig:fifo_srtf_sl_drl}
\end{minipage}%
\vspace{-4mm}
\end{figure}

\begin{figure}[t]
\centering
\begin{minipage}[t]{0.495\linewidth}
\centering
\includegraphics[height=1.2in]{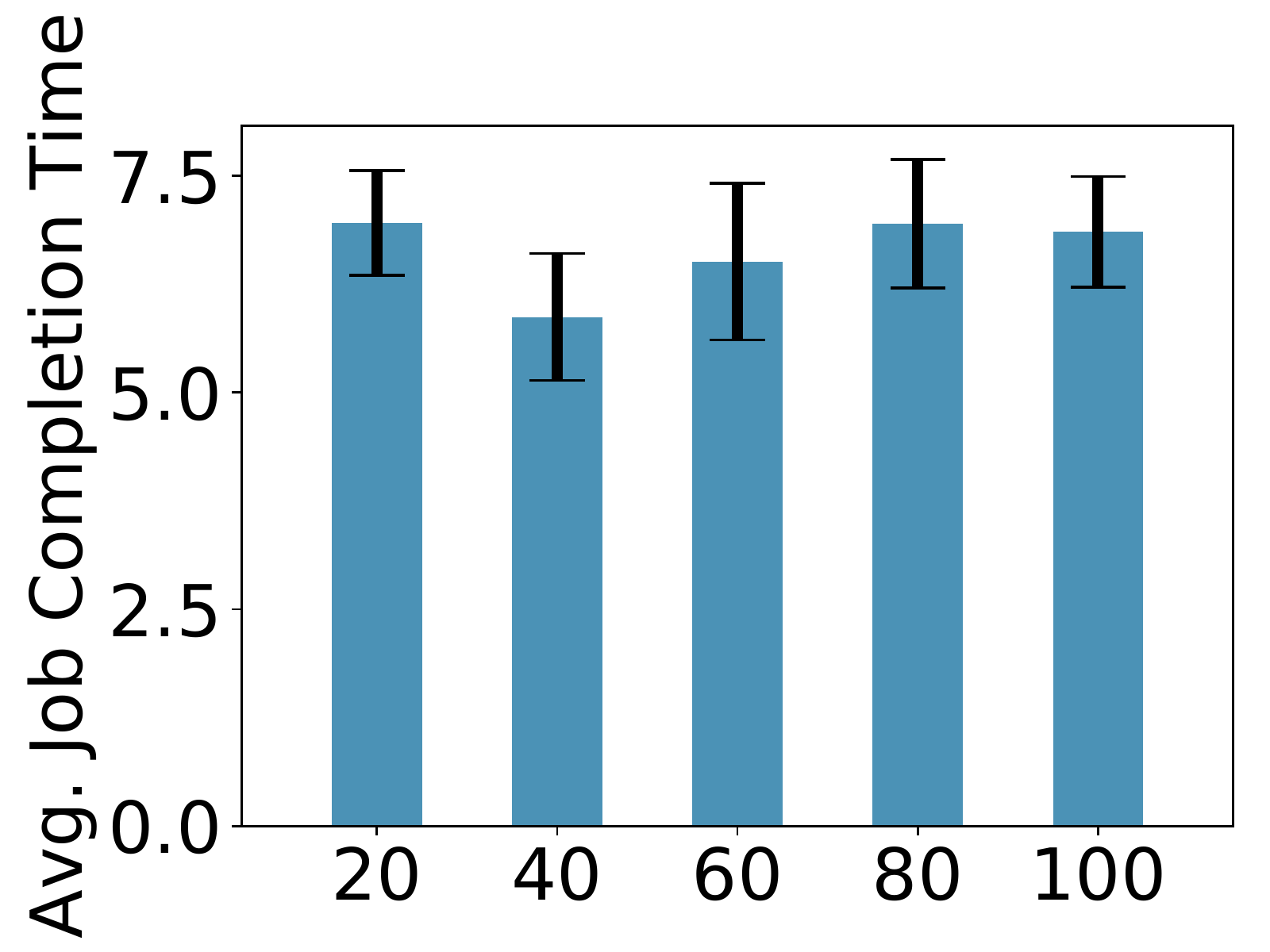}
\captionof{figure}{Concurrent\\ job number}
\label{fig:concurrent_job_number}
\end{minipage}%
\begin{minipage}[t]{0.495\linewidth}
\centering
\includegraphics[height=1.2in]{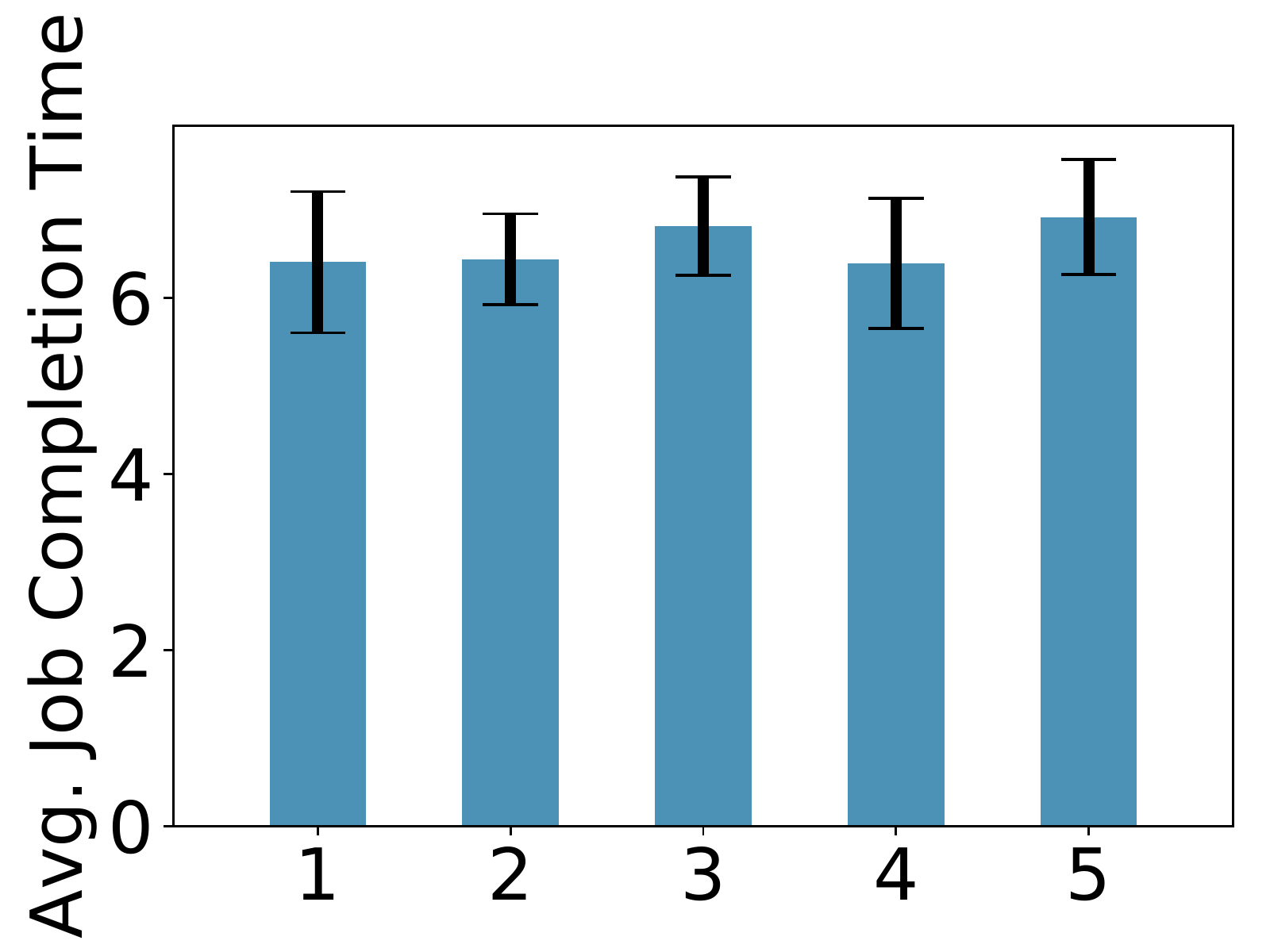}
\captionof{figure}{Varying cluster number} 
\label{fig:vary_cluster_num}
\end{minipage}%
\vspace{-4mm}
\end{figure}

{\noindent\textbf{Unseen job types.}}
We investigate whether \DL2~can adapt to jobs training new models. We train the neural network using the first four categories of models (Table~\ref{table:job_details}) in the supervised learning phase and the first 1000 steps of the online RL phase. At step 1000 and step 2000 of the RL phase (\ie, the red dots in Fig.~\ref{fig:unseen_models}), we submit jobs training two new categories of models. In the case of the ``ideal'' baseline, we train the NN using all categories of jobs in Table~\ref{table:job_details} from the beginning. Fig.~\ref{fig:unseen_models} shows the average job completion time achieved using the trained NN at each time respectively, for decision making over the validation dataset. \DL2~gradually achieves the same performance as the ``ideal'' baseline, showing its capability to handle new types of DL jobs coming on the go.


{\noindent\textbf{Other scheduling strategies for supervised learning.}}
We change the default DRF used in supervised learning of \DL2~to two other heuristics, First-In-First-Out (FIFO) and Shortest-Remaining-Time-First (SRTF). Fig.~\ref{fig:fifo_srtf_sl_drl} shows average job performance when \DL2~uses each of these strategies in its supervised learning phase, when the NN trained only using supervised learning, or using both supervised learning and online RL, is evaluated on the validation dataset. 
In both cases, the performance is significantly improved with \DL2, beyond what the existing scheduling strategy in the cluster can achieve ($41.3\%$ speedup in the case of SRTF).

{\noindent\textbf{Concurrent job number.}} We investigate how the maximal number of concurrent jobs to schedule in a time slot, $J$ specified in the NN input, affects the performance of \DL2~when applying the trained NN (after supervised learning and reinforcement learning) on the validation dataset. The maximal number of uncompleted jobs in all time slots is around 40; when the concurrent job number is larger than $J$, we schedule the jobs in batches of $J$ jobs, according to their arrival sequence. In Fig.~\ref{fig:concurrent_job_number}, 
 we observe that the performance suffers when $J$ is small, possibly because the NN is not trained on a global view when jobs are fed into the NN in batches in each time slot. Setting $J$ to be large enough to accommodate the maximal number of concurrent jobs gives better results.


\begin{table}[t]
\vspace{2mm}
\centering
\footnotesize
\caption{Effectiveness of Training Techniques}
\label{table:vary_design_choices}
\begin{tabular}{|c|c|c|}
\hline
Without							& Avg. Job Completion Time & Slowdown (\%)						\\
\hline
-										& $5.724\pm0.844$	&	0			\\
\hline
Actor-critic						& $6.929\pm0.477$	&	21.1			\\
\hline
Exploration						& $7.372\pm0.548$	&	28.8			\\
\hline
Experience replay			& $7.988\pm0.102$	&	39.6			\\
\hline
\end{tabular}
\vspace{-4mm}
\end{table}

\subsection{Training Design}
{\noindent\textbf{SL loss function.}} We evaluate three common loss functions for supervised learning, \ie, Mean Square, Cross Entropy (the default) and Absolute Difference~\cite{tflearn_objectives}. 
 We observe similar performance with these loss functions, while adopting Cross Entropy achieves the best performance. This is because Mean Square or Absolute Difference emphasize incorrect or suboptimal output, while only the correct or optimal output contributes to the loss when using Cross Entropy.

{\noindent\textbf{Reward function.}} We evaluate another reward function with \DL2, which sets the reward of each action (that adds some worker/PS to a job) as the normalized number of epochs trained by the job in the time slot. We find that its performance is $29.1\%$ worse. Our default reward function considers all jobs' progress, enabling the policy network to learn to schedule from a global perspective.

{\noindent\textbf{Actor-critic.}} To see how the actor-critic algorithm affects training, we remove the value network but only train the policy network. As widely adopted in RL community, we use the exponential moving average of rewards as a baseline in place of the output of the value network in gradient computation of the policy network. 
 As shown in Table~\ref{table:vary_design_choices}, with the value network, the performance is $21.1\%$ better. This is because the average reward is not always an effective baseline; in some cases, even the optimal action leads to a lower reward than the average reward.
 
{\noindent\textbf{Job-aware exploration.}} We examine how exploration contributes to the performance. From Table~\ref{table:vary_design_choices}, we see that without exploration the performance is $28.8\%$ worse, as online RL is stuck in a local optimal policy. 

{\noindent\textbf{Experience replay.}} 
We disable experience replay and see how performance changes. Table~\ref{table:vary_design_choices} shows that the average job completion time is degraded by 39.6\%, indicating that experience replay is critical for training. 

{\noindent\textbf{Federated training.}}
Federated training enables multiple clusters to learn a global \DL2~model collaboratively. We study how the number of clusters affects the policy training, by implementing the A3C~ \cite{mnih2016asynchronous} algorithm, which trains a global policy NN using multiple \DL2~schedulers with different training datasets, each for one cluster. Fig.~\ref{fig:vary_cluster_num} shows that the global performance remains stable when we increase the number of clusters. We have also observed that with more clusters, the policy NN converges much faster due to the use of more training datasets: if there are $x$ clusters, the NN converges almost $x$ times faster. The preliminary result also suggests the possibility of dividing a single massive cluster into loosely coupled sub-clusters where each runs a \DL2~scheduler for resource allocation, if scalability issue arises.

\vspace{-4mm}
\section{Discussion and Future Directions}
\label{sec:discussions}

{\noindent\textbf{More scheduling features.}}
Besides minimizing average job completion time, \DL2~can implement other scheduling features by adjusting the learning objective. For example, we can incorporate resource fairness by adding a quantified fairness term in the reward function.

{\noindent\textbf{All-reduce architecture.}}
All-reduce architecture~\cite{watcharapichat2016ako}, where workers train model replicas and exchange updated model parameters directly with each other, is supported in Caffe2~\cite{Caffe2}, CNTK~\cite{cntk}, etc. Though this paper focuses on the PS architecture, \DL2~can readily handle jobs using all-reduce architecture with minor modification of input state and action space of its NN, \eg, removing the elements related to PSs.

{\noindent\textbf{Job placement.}}
While we use the default placement policy in this work, the placement of workers and PSs can potentially be decided by RL too. Using one NN to produce both resource allocation and placement decisions is challenging, mainly because of the significantly larger action space. RL using a hierarchical NN model~\cite{mirhoseini18hierarchical} might be useful in making resource allocation and placement decisions in a hierarchical fashion.

{\noindent\textbf{Practical deployment.}} In practical deployment, the following two issues may need to be considered: (1) adversarial attacks that fool a neural network with malicious input; (2) neural network monitoring that detects exceptional scheduling. These are interesting directions to explore, with progress in security research and more in-depth understanding of neural networks.
\section{Related Work}
\label{related_DL_in_network}

\vspace{1mm}
\noindent{\bf Deep reinforcement learning in system research.}
A number of recent studies use DRL for resource allocation, device placement, and video streaming. 
Mao {\em et al.}~\cite{mao2016resource} and Chen {\em et al.}~\cite{chen2017deep} use DRL for job scheduling in cloud clusters, to minimize average job slowdown. 
Their NNs select the jobs (single-task jobs) to run with static resource allocation. 
The NNs are trained offline: multiple job arrival sequences are used as training examples; each example is repeatedly trained for multiple epochs. 
Mao {\em et al.}~\cite{mao2018learning}\cite{mao2019learning} learn an NN to schedule graph-based parallel jobs as in Spark, in terms of parallelism level and execution order of tasks in the jobs, using offline training. Adjustment of resources during job execution is not in the scope of the above studies.

Mirhoseini {\em et al.}~\cite{mirhoseini2017device}\cite{mirhoseini18hierarchical} use DRL to optimize placement of a computation graph, to minimize running time of an individual TensorFlow job.  
Xu {\em et al.}~\cite{xu2018experience} use DRL to select routing paths between network nodes for traffic engineering. 
Mao {\em et al.}~\cite{mao2017neural} dynamically decide video streaming rates in an adaptive streaming system with DRL. 
All these studies resort to offline RL training, using data generated by analytical models or simulators. In contrast, we use offline supervised learning to prepare our NN and then online RL to further improve the NN. 

\vspace{1mm}
\noindent{\bf ML cluster scheduling.}
SLAQ~\cite{zhang2017slaq} adopts online fitting to estimate the training loss of convex algorithms, for scheduling jobs training classical ML models. Dorm~\cite{sun2017towards} uses a utilization-fairness optimizer to schedule ML jobs. These work do not focus on distributed ML jobs using the parameter server architecture. Optimus~\cite{peng2018optimus} proposes a dynamic resource scheduler based on online-fitted resource-performance models. Bao {\em et al.}~\cite{bao2018online} design an online scheduling algorithm for DL jobs. These studies rely on detailed modeling of DL jobs and simplified assumptions in their design. Gandiva~\cite{Gandiva} exploits intra-job predictability to time-slice GPUs efficiently across multiple jobs, and dynamically migrate jobs to better-fit GPUs. They do not consider resource allocation adjustment; Resource allocation with GPU sharing will be an intriguing future direction to explore.

\section{Conclusions}

We present \DL2, a DL-driven scheduler for DL clusters, which expedites job completion globally with efficient resource utilization. \DL2~starts from offline supervised learning, to ensure basic scheduling performance comparable to the existing cluster scheduler, and then runs in the live DL cluster to make online scheduling decisions, while improving its policy through reinforcement learning using live feedback. Our testbed experiments and large-scale trace-driven simulation verify \DL2's low scaling overhead, generality in various scenarios and outperformance over hand-crafted heuristics.

\bibliographystyle{abbrv}
\bibliography{reference}

\end{document}